\documentclass[sigconf,dvipsnames]{acmart}

\AtBeginDocument{%
  \providecommand\BibTeX{{%
    \normalfont B\kern-0.5em{\scshape i\kern-0.25em b}\kern-0.8em\TeX}}}
    
\copyrightyear{2024}
\acmYear{2024}
\setcopyright{acmlicensed}
\acmConference[MM '24]{Proceedings of the 32nd ACM International Conference on Multimedia}{October 28-November 1, 2024}{Melbourne, VIC, Australia}
\acmBooktitle{Proceedings of the 32nd ACM International Conference on Multimedia (MM '24), October 28-November 1, 2024, Melbourne, VIC, Australia}
\acmDOI{10.1145/3664647.3680664}
\acmISBN{979-8-4007-0686-8/24/10}

\usepackage{microtype}
\usepackage{graphicx}
\usepackage{amsmath}
\usepackage{bbm}
\usepackage{mathtools}
\usepackage[ruled,vlined]{algorithm2e}
\usepackage{cleveref}
\usepackage{subcaption}
\usepackage{booktabs} 
\usepackage{multirow}
\usepackage{tikz}
\usepackage{xcolor}
\usepackage{colortbl}
\def\approxprop{%
  \def\p{%
    \setbox0=\vbox{\hbox{$\propto$}}%
    \ht0=0.6ex \box0 }%
  \def\s{%
    \vbox{\hbox{$\sim$}}%
  }%
  \mathrel{\raisebox{0.7ex}{%
      \mbox{$\underset{\s}{\p}$}%
    }}%
}

\usepackage{amsmath,amsfonts,bm}

\def\eqref#1{equation~\ref{#1}}

\def\1{\bm{1}}

\def\ry{{\textnormal{y}}}

\def\rvx{{\mathbf{x}}}

\def\rvz{{\mathbf{z}}}

\def\vp{{\bm{p}}}

\def\vx{{\bm{x}}}

\def\vz{{\bm{z}}}

\DeclareMathAlphabet{\mathsfit}{\encodingdefault}{\sfdefault}{m}{sl}
\SetMathAlphabet{\mathsfit}{bold}{\encodingdefault}{\sfdefault}{bx}{n}

\newcommand{\softmax}{\mathrm{softmax}}

\begin{document}

\title{CLIPCleaner: Cleaning Noisy Labels with CLIP} 


\author{Chen Feng}
\affiliation{%
  \institution{Queen Mary University of London}
  \city{London}
  \country{UK}}
\email{chen.feng@qmul.ac.uk}

\author{Georgios Tzimiropoulos}
\affiliation{%
  \institution{Queen Mary University of London}
  \city{London}
  \country{UK}}
\email{g.tzimiropoulos@qmul.ac.uk}

\author{Ioannis Patras}
\affiliation{%
  \institution{Queen Mary University of London}
  \city{London}
  \country{UK}}
\email{i.patras@qmul.ac.uk}

\begin{abstract}
Learning with Noisy labels (LNL) poses a significant challenge for the Machine Learning community. Some of the most widely used approaches that select as clean samples for which the model itself (the in-training model) has high confidence, e.g., `small loss', can suffer from the so called `self-confirmation' bias. This bias arises because the in-training model, is at least partially trained on the noisy labels. Furthermore, in the classification case, an additional challenge arises because some of the label noise is between classes that are visually very similar (`hard noise'). 
This paper addresses these challenges by proposing a method (\textit{CLIPCleaner}) that leverages CLIP, a powerful Vision-Language (VL) model for constructing a zero-shot classifier for efficient, offline, clean sample selection. 
This has the advantage that the sample selection is decoupled from the in-training model and that the sample selection is aware of the semantic and visual similarities between the classes due to the way that CLIP is trained. We provide theoretical justifications and empirical evidence to demonstrate the advantages of CLIP for LNL compared to conventional pre-trained models.
Compared to current methods that combine iterative sample selection with various techniques, \textit{CLIPCleaner} offers a simple, single-step approach that achieves competitive or superior performance on benchmark datasets. To the best of our knowledge, this is the first time a VL model has been used for sample selection to address the problem of Learning with Noisy Labels (LNL), highlighting their potential in the domain.
\end{abstract}


\begin{teaserfigure}
\centering
   \includegraphics[width=1\textwidth]{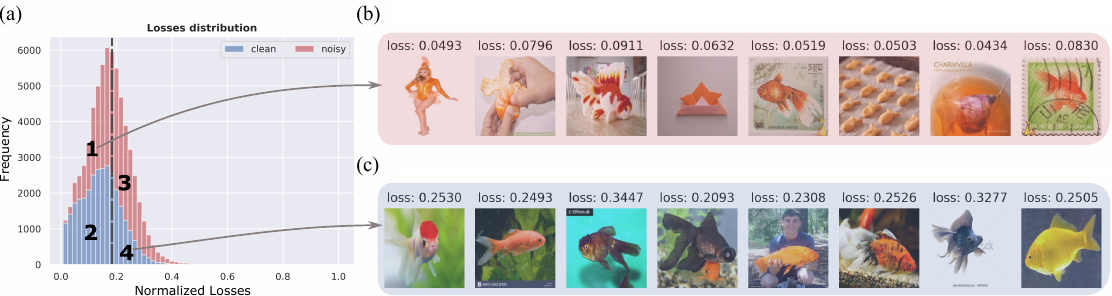}
    \caption{The normalized losses distribution of WebVision dataset after one epoch warm-up training, i.e., training with whole dataset and cross-entropy loss. In (a), `clean'/`noisy' denotes samples been identified as clean/noisy by \textit{CLIPCleaner} while the `gray vertical line' denotes the sample selection boundary induced by `small-loss' mechanism. We show some example images on \textit{part \textsc{1}} in (b) and \textit{part \textsc{4}} in (c) which represents samples identified as `clean' by `small-loss' while rejected by \textit{CLIPCleaner} and vice versa. 
    For example, in (b) we can find that many images with small losses due to its similar color or textures to `tench' class, thus been wrongly identified as `clean' by `small-loss' but been correctly rejected by \textit{CLIPCleaner}.
    }
    \label{fig:CLIP_intuition}
\end{teaserfigure}

\maketitle

\section{Introduction}
Over the past two decades, deep neural networks have demonstrated exceptional success in various vision tasks, partly due to the existence of accurately labelled, large-scale datasets such as ImageNet-1K. However, collecting high-quality labels for such datasets is generally time-consuming and labour-intensive. 
Noisy labels, stemming from human error, ambiguity in labelling criteria, or inherent noise in data collection processes, introduce a critical challenge that traditional learning algorithms must deal with.

To learn with noisy labels (LNL), various methods have been proposed. Some methods aim to develop robust loss functions~\cite{generalized_cross_entropy,symmetric_cross_entropy, ye2024active, mae, feng2021can, ma2020normalized, wang2021learning, zhou2021asymmetric} or model the labeling error patterns with a label transition matrix~\cite{noiseadaptation,loss_correction,loss_correction2, li2022estimating, yao2020dual, xia2022extendedT}. However, these methods are often sub-optimal in dealing with high noise ratios and complicated noise patterns.

More recently, methods based on sample selection~\cite{sun2022pnp_selection,wei2022self_selection,wang2022promix_selection,karim2022unicon_selection,patel2023adaptive_selection,zhang2021dualgraph_selection, park2024robust, kim2023crosssplit, kim2024neural} that aim to identify samples with clean labels have become perhaps the dominant paradigm.
Among them, the most common sample selection strategies are the `small-loss' mechanism motivated by the fact that the model tends to fit clean samples earlier than noisy samples in the training process -- this  results in relatively smaller losses for the clean samples.
Following this, most of methods focus primarily on further improving such sample selection mechanisms. This includes different variants of the `small-loss' strategy~\cite{dividemix,cnlcu,lossmodellingbmm}, and utilizing kNN~\cite{deep-knn,moit,ssr} or graph models~\cite{topofilter,ngc} based on the samples' feature space for sample selection. However, these methods are inherently affected by the label noise, as losses, or the features used for sample selections are extracted from the model that is being trained (i.e., the in-training model) -- this leads to the infamous `self-confirmation' bias. Some methods~\cite{coteaching,coteaching+} attempt to \textit{alleviate} `self-confirmation' bias through model co-training, but this approach introduces additional computational overhead. 
Moreover, these methods solely rely on the visual information within the images, and therefore have difficulty dealing with `hard noise', that is labelling errors between classes with high visual similarity.

To address the aforementioned issue, we propose a novel method, namely \textit{CLIPCleaner}, that leverages the popular visual-language model CLIP~\cite{clip} for sample selection. Specifically, we propose using a CLIP-based zero-shot classifier with descriptive class prompts that are generated automatically using a Large Language Model
for sample selection. Given that CLIP is trained with massive vision-language pairs, this leads to a sample selection scheme that has two advantages: 
1. the sample selection is aware of visual and semantic similarities between the classes and therefore compensates for biases that may arise from relying solely on visual information for sample selection~(\cref{fig:CLIP_intuition}); 2. the sample selection is independent of the in-training model, and therefore immune to the influence of noisy labels and the `self-confirmation' bias.
\textbf{To the best of our knowledge, we are the first to employ a large-scale vision-language model, particularly leveraging its language modality, for sample selection.} 

Furthermore, we introduce a very simple semi-supervised learning method tailored for noisy datasets without common advanced modules such as co-training or multi-task training, namely \textit{MixFix}. The proposed semi-supervised method, gradually introduces more clean samples and re-labels noisy samples to expand the initial clean subset selected by \textit{CLIPCleaner}. 
Let us note that in the proposed scheme, the in-training model, i.e., the final classifier, is different from the VL model that is used for sample selection. More specifically, unlike common transfer learning techniques such as model fine-tuning~\cite{gao2021clipadaptor}, knowledge distillation~\cite{wang2022cliptd}, and prompt-based learning~\cite{CoOp,PLOT}, we adhere to using CLIP solely for sample selection and refrain from training/fine-tuning it. This has the distinct advantage that the proposed scheme allows for computationally, or parameter-wise light in-training model, and allows the use as sample selectors of VL models to which one does not necessarily have full access. 

We demonstrate the effectiveness and advantages of the proposed method both theoretically and empirically. Despite its simplicity, our method achieves competitive and superior performance on various datasets, including CIFAR10/CIFAR100 with synthetic noise~(symmetric, asymmetric, and instance-dependent), as well as real-world noisy datasets like Red Mini-ImageNet, WebVision, Clothing1M, and ANIMAL-10N.

\section{Related works}
\paragraph{Sample selection for learning with noisy labels}
Most sample selection methods usually rely on model classifiers, such as the widely-applied `small-loss' mechanism~\cite{lossmodellingbmm,dividemix,coteaching,mentornet} or model predictions~\cite{selfie,whentohow,pencil}.
More recent works focus on further improving the sample selection quality by modelling the loss with markov process~\cite{cnlcu} or dynamically selecting samples with multiple metrics~\cite{zhou2020robust}. 
In addition,
some works try to utilize the feature representations for sample selection. \citet{topofilter} and \citet{ngc} try to build a kNN graph and identify clean samples through connected sub-graphs,  while \citet{ssr, noisebox, moit} propose to utilize a kNN in feature space to alleviate the effect of noisy labels. Some recent methods involving contrastive learning also identify clean sample pairs based on neighbourhood relationships in the feature space~\cite{selcl} or fit Gaussian distributions to model the clean distribution~\cite{tcl}. However, these methods remain unstable and prone to `self-confirmation' bias, especially in high-ratio noise scenarios, due to their intrinsic reliance on the in-training model based on noisy datasets.

\paragraph{Utilization of auxiliary model}
The utilization of an auxiliary noise-free model is reasonable and straightforward for LNL. Related to us, some methods also try to use pre-trained noise-free models for learning with noisy labels. \citet{c2d,cheng2021demystifying} propose to utilize self-supervised learning~\cite{simclr, moco, feng2022adaptive, Feng_2023_CVPR, Sun_2024_CVPR, Gao_2024_WACV,Metaxas_2023_CVPR, Gao_2024_CVPR} since it can learn good representations in the label-free case. \citet{deep-knn} utilize the pre-logit space of the pre-trained model along with the kNN classifier for sample selection. \citet{detecttodenoise} follow the same idea and also involve CLIP, but they only utilize its vision encoder as a common pre-trained encoder without utilizing the language encoder. In this work, we emphasize that language modality is critical as a supplementary modality and show the unique advantage of VL models for sample selection, both theoretically and empirically.

\begin{figure*}[htbp]
    \centering
    \includegraphics[width=0.7\linewidth]{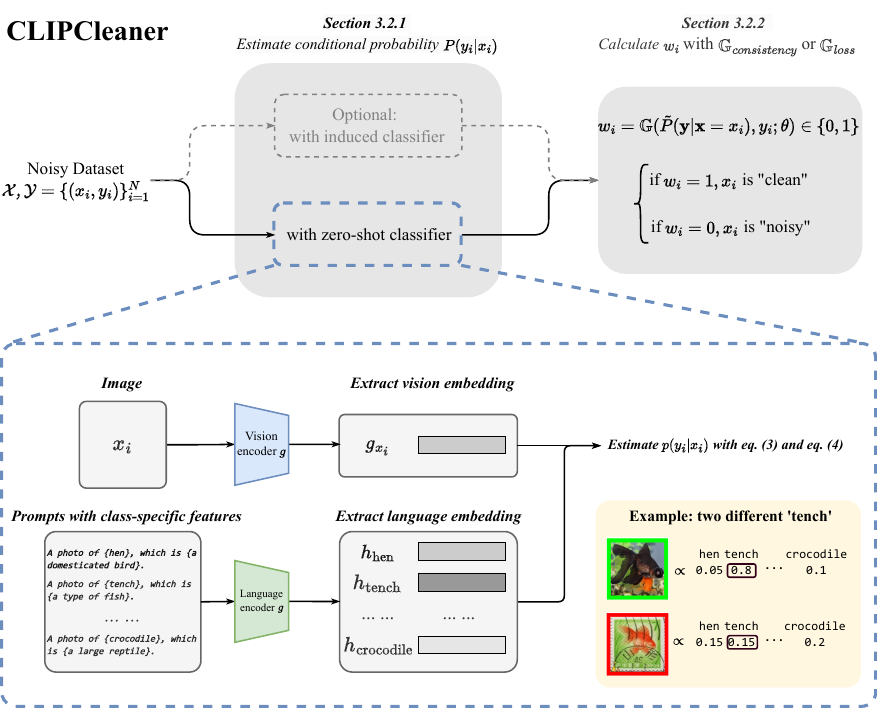}
    \caption{Workflow of \textit{CLIPCleaner}. We highlight the sections corresponding to the two main steps of \textit{CLIPCleaner}, and particularly visualize the intuition of the probability estimation step based on the CLIP zero-shot classifier.}
    \label{fig:workflow}
\end{figure*}

\section{Method}
\label{sec:meth}
In \cref{sec:weight}, we cast the learning with noisy labels problem in a formulation that covers mainstream sample selection methods. In \cref{sec:meth_robustclip}, we elaborate our sample selection method, namely \textit{CLIPCleaner}. In \cref{sec:semi}, we introduce our semi-supervised learning method, namely \textit{MixFix}. In \cref{sec:theory} , we theoretically analyze the unique advantage of using CLIP for sample selection over common pretrained models. In \cref{sec:extra_discussion}, we provide further discussions on the topics of sample selection and the usage of the CLIP model.

\subsection{Revisiting sample selection for LNL}\label{sec:weight}
Given a dataset of training samples $\{\vx_i,y_i\}_{i=1}^{N}$ \textit{i.i.d} sampled from a noisy joint distribution $P^n(\rvx,\ry)$, 
the goal is to learn a classifier $f:\rvx\rightarrow \ry$ that can accurately predict the true labels $y$ for new, unseen clean examples. Let us denote the clean (but unknown) joint distribution as $P(\rvx,\ry)$.
Sample selection methods aim to identify those samples with (possibly) clean labels.
Let us also denote the sample selection results as $\{w_i \in \{0, 1\}\}_{i=1}^N$ with $w_i = 1$ or $0$ representing sample $\vx_i$ been selected or not. 
For specific sample $\vx_i$, here, allowing us to propose a concise form to represent most existing sample selection methods:
\begin{equation}\label{eq:weight_estimation}
w_i = \mathbb{G}(\Tilde{P}(\ry|\rvx = \vx_i), y_i; \theta) \in \{0,1\}.
\end{equation}
We define as $\Tilde{P}(\ry|\rvx = \vx_i)$ an estimation of the clean conditional probability $P(\ry|\rvx = \vx_i)$, and abbreviate here as $\mathbb{G}$ a specific selection mechanism, with its hyperparameter as $\theta$. 
Intuitively speaking, we conceptualize a sample selection method into two steps: firstly estimating clean conditional probability $\Tilde{P}(\ry|\rvx = \vx_i)$ for sample $\vx_i$. Then, applying $\mathbb{G}$ to compare $\Tilde{P}(\ry|\rvx = \vx_i)$ with the annotated label $y_i$, to decide/measure if the annotated label is (likely) clean or not\footnote{Please note, there are indeed some methods such as TopoFilter~\cite{topofilter} and FINE~\cite{kim2021fine} relying on graph models or eigenvectors rather than probability estimations for sample selection. With \cref{eq:weight_estimation} we are not attempting to cover all possible sample selection mechanisms but to motivate our proposed method.}. 

However, most sample selection methods inherently and inevitably lead to the `self-confirmation' bias as they commonly (more or less) rely on the in-training model $f$ in estimating the conditional probability: $\Tilde{P}(\ry|\rvx = \vx_i) = P_f(\ry|\rvx = \vx_i)$. 
To fully avoid such `self-confirmation' bias - the reliance of sample selection on in-training model $f$, utilizing another pre-trained classifier naturally fits. Specifically, in this work, we consider to utilize the CLIP model for sample selection. 

\subsection{CLIPCleaner: sample selection with vision-language models}
\label{sec:meth_robustclip}

\subsubsection{Preliminary on CLIP}
We first briefly introduce the CLIP model~\cite{clip}, which is currently one of the most prevalent vision-language models. CLIP aims to learn from a dataset of image-text pairs, denoted as ${(\vx'_i,\vz_i)}_{i=1}^{M}$, which is \textit{i.i.d.} sampled from a hidden joint distribution $Q(\rvx,\rvz)$. Specifically, we consider $Q(\rvx,\rvz)$ as the marginalization of $Q(\rvx, \ry, \rvz)$ for ease of later analysis. We denote as $\vx'$ the images in CLIP training dataset to discriminate from above in-question noisy dataset, and $\vz$ the corresponding text descriptions. Then, we have below as CLIP training loss:
\begin{align} \label{eq:clip_loss}
\begin{split}
    L(\vx'_i,\vz_i;g, h) = \frac{1}{2} \big(-\log \frac{\exp(g(\vx'_i)^T h(\vz_i))}{\sum_{j=1}^M \exp(g(\vx'_i)^T h(\vz_j))} \\-\log \frac{\exp(g(\vx'_i)^T h(\vz_i))}{\sum_{j=1}^M \exp(g(\vx'_j)^T h(\vz_i))}\big).
\end{split}
\end{align}
Here, $g$ and $h$ denote the vision and language encoder, respectively. Intuitively, the CLIP model tries to maximize the correspondence between related image-text pairs.

\subsubsection{Estimate $P(\ry|\rvx = \vx_i)$ with CLIP zero-shot classifier}
Due to its multimodal nature, CLIP naturally possesses the ability for zero-shot classification. As a relatively new technology for the LNL community, here we revisit CLIP's zero-shot classification from a probabilistic perspective, which will also serve as our method for estimating true conditional probabilities with CLIP.

Let us recall $\vx, y, \vz$ as the image, label and text respectively. Firstly, we assume $y \,\bot\, \vx \mid \vz$; intuitively, the semantic label $y_i$ can be independently generated based on a decent image description $\vz_i$ alone for each image $\vx_i$. For zero-shot classification, we have:
\begin{align}\label{eq:zeroshot}
\begin{split}
    P_{zeroshot}(\ry=y_i|\rvx=\vx_i) &= \int Q(\ry=y_i|\rvz=\vz_i)Q(\rvz=\vz_i|\rvx=\vx_i)dz \\&\propto \int Q(\ry=y_i|\rvz=\vz_i)Q(\rvz=\vz_i,\rvx=\vx_i)dz.   
\end{split}
\end{align}
To calculate above integral analytically is often hard; Practically, we tend to estimate $P_{zeroshot}(\ry=y_i|\rvx=\vx_i)$ by sampling $\vz_i$, if $Q(\ry=y_i|\rvz=\vz_i)$ and $Q(\rvz=\vz_i,\rvx=\vx_i)$ is known. Firstly, according to the training loss used by CLIP, we know that\footnote{Please refer to \textsc{Supplementary E} for full derivation.}:
$$Q(\rvz=\vz_i,\rvx=\vx_i) \propto \exp(g(\vx_i)^T h(\vz_i))).$$
Still, \textcolor{black}{$Q(\ry=y_i|\rvz=\vz_i)$} remains unknown. Original CLIP designs a single prompt as `\texttt{A photo of {class name of $y_i$}.}', implicitly assuming that: $$Q(\ry=y_i|\rvz=\text{`\texttt{A photo of {class name of $y_i$}.}'}) \approx 1.$$ Then, with a single prompt we can easily sample a single $\vz_i$ to estimate $P_{zeroshot}(\ry=y_i|\rvx=\vx_i)$ according to \cref{eq:zeroshot}.
Moreover, it is plausible that with more high-quality samplings of $\vz_i$ instead of only utilizing one single prompt the estimation would be better. In this work, we apply below template to generate multiple prompts $\{\mathcal{P}_j\}_{j=1}^J$ using class-specific features such as the unique color or habitat of different animal species in an animal classification task\footnote{Please refer to \textsc{Supplementary B} for more details about prompts generation.}:
\begin{center}
    $\mathcal{P}_j=$`\texttt{A photo of \{class name of $y_i$\}, which is/has \{class-specific feature $j$ of class $y_i$\}.}'
\end{center}
Then, we can similarly estimate $P_{zeroshot}(\ry=y_i|\rvx=\vx_i)$ with above prompts as below:
\begin{align}\label{eq:estimation_zeroshot}
\begin{array}{lc}
        P_{zeroshot}(\ry=y_i|\rvx=\vx_i) \approxprop \sum_{j=1}^J\Tilde{Q}(\rvz=\mathcal{P}_j ,\rvx=\vx_i).
\end{array}
\end{align}

\subsubsection{Calculate $w_i$ with specific $\mathbb{G}$}
With the above estimated conditional probability $\Tilde{P}(\ry|\rvx = \vx_i) = P_{zeroshot}(\ry=y_i|\rvx=\vx_i)$, we can apply any applicable strategy for sample selection as depicted in \cref{eq:weight_estimation}. As exploration on more advanced sample selection strategy $\mathbb{G}$ is not the focus in this paper, we consider two simple sample selection strategies below.  

Firstly, we consider a \textit{consistency-based selector} - compute sample's consistency metric (defined as the ratio of the probability of noisy label class to the highest class probability) and identify samples with high consistency as clean: 
\begin{equation}\label{eq:consselection}
    \mathbb{G}_{consistency} = \mathbb{I}(\frac{\Tilde{P}(\ry = y_i|\rvx = \vx_i)}{\max_k\Tilde{P}(\ry = k|\rvx = \vx_i)} \geq \theta_{consistency}).
\end{equation}
Here, $\mathbb{I}$ is the indicator function,  $\theta_{consistency}$ is the manually-defined threshold, often as 1 by default.

Denoting as $\{\Tilde{P}(\ry|\rvx = \vx_i)\}_{i=1}^N$ the estimated probabilities for the training dataset, we also consider the widely-applied \textit{loss-based sample selector} - computing the sample's cross-entropy loss~($\{-\log \Tilde{P}(\ry = y_i|\rvx = \vx_i)\}_{i=1}^N$) and then dividing the dataset into two parts based on a Gaussian Mixture Model (GMM), with the part having smaller losses designated as clean samples:
\begin{equation}\label{eq:lossselection}
    \mathbb{G}_{loss} = \mathbb{I}(\mathbb{P}(-\log \Tilde{P}(\ry = y_i|\rvx = \vx_i) \in \mathtt{GMM}_{small}) \geq \theta_{loss}).
\end{equation}
Due to the possible class imbalances and the various semantic diversity of different classes, slightly different than the common approach utilizing a single GMM, we model the losses of samples from each class by a separate GMM model\footnote{Please refer to \textsc{Supplementary D} for specific ablations.}. Here, $\theta_{loss}$ is also the manually-defined threshold, often as 0.5 by default.

\subsection{MixFix: Efficient semi-supervised training by absorbing and relabelling}\label{sec:semi}
With selected subset only, \textit{CLIPCleaner} can be utilized along with any existing methods - see \textsc{Supplementary C} for results of utilizing \textit{CLIPCleaner} with DivideMix~\cite{dividemix}. 
However, the state-of-the-art methods often involve multiple modules, such as iterative sample selection and model training~\cite{dividemix,ssr,tcl}, model co-training~\cite{coteaching, coteaching+}, and multi-task contrastive learning~\cite{moit, selcl}. While these modules can be effective, they introduce extra complexity into the learning process, requiring careful coordination and tuning. To streamline the process and enhance efficiency, we aim to avoid intricate methodologies and propose a simple semi-supervised learning method for noisy datasets --- namely \textit{MixFix}. 

Let us denote the selected subset and non-selected subset as $\mathcal{X}_c, \mathcal{Y}_c$ and $\mathcal{X}_n, \mathcal{Y}_n$. 
Motivated by FixMatch~\cite{fixmatch}, we also inspect in unlabeled subset ($\mathcal{X}_n, \mathcal{Y}_n$) each sample's current prediction $\vp_i$ based on the in-training model $f$:

\begin{equation} \label{eq:MixFix}
    (w_i, y_i) =\left\{ \\
\begin{aligned}
    &(0, y_i), \ \text{if } p_m < \theta_r \text{ and } p_m < \theta'_r &~\text{\texttt{*Drop*}}\\
    &(1, y_i), \ \text{if } p_m > \theta_r \text{ and } y_i = y_m  &~\text{\texttt{*Absorb*}}\\
    &(1, y_m), \ \text{if } p_m > \theta'_r \text{ and } y_i \neq y_m &~\text{\texttt{*Relabel*}}\\
  \end{aligned}
\right.
\end{equation}
Here we denote as $p_m \triangleq \max_l \vp_i(l)$ and $y_m \triangleq \arg\max_l \vp_i(l)$. Please note the difference of $\vp_i$ here with our previous estimated probabilities for sample selection. Intuitively, we `\texttt{absorb}' more clean samples~($y_i = y_m$) (not been selected in the sample selection step) and `\texttt{relabel}' noisy samples~($y_i \neq y_m$) with different thresholds ($\theta_r$ and $\theta'_r$) in non-selected subset, and progressively append it to initial selected subset to form a dynamic larger training set $\mathcal{X}_t, \mathcal{Y}_t$. Different from FixMatch~\cite{fixmatch} using one threshold for all samples, we typically set $\theta_r \leq \theta'_r$. This allows us to fully leverage noisy labels to distinguish between the `\texttt{absorb}' and `\texttt{relabel}' processes.
Then, we apply a common cross-entropy loss for training with this expanded training set $\mathcal{X}_t, \mathcal{Y}_t$.

\subsection{Theoretical justification of \textit{CLIPCleaner}}
\label{sec:theory}
Ignoring the language modality and treating the CLIP model as an ordinary pre-trained model, we can also leverage its vision encoder $g$ solely along with the interested noisy dataset ${(\vx_i,y_i)}_{i=1}^{N}$ to induce a new classifier $f'$ (to discriminate it with the model $f$ in \cref{sec:weight}) for estimating the clean conditional probability in \cref{eq:weight_estimation}. 
For example, we can simply freeze the weights of $g$, use it as a fixed feature encoder, and train a linear classifier $f'$ upon it based on the interested noisy dataset. Then the predicted logits after softmax normalization can be used as an estimate of $P(\ry|\rvx=\vx_i)$:
\begin{equation}\label{eq:estimation_noisy}
          P_{induced}(\ry|\rvx=\vx_i) = \softmax(f'(g(\vx_i))).
\end{equation}
\textit{An immediate question is: how does the zero-shot classifier~(\cref{eq:estimation_zeroshot}) compare to the induced classifier here~(\cref{eq:estimation_noisy}) in estimating the clean conditional probability?} In fact, the induced classifier can be based on any visual pre-trained model. If such easily-induced classifier demonstrates performance comparable to or even better than the zero-shot classifier, then we have no motivation to specifically adopt the CLIP model for sample selection.
To this end, we conduct a theoretical analysis and compare the estimated $\Tilde{P}(\ry|\rvx=\vx_i)$ in both options with the true/unknown $P(\ry|\rvx=\vx_i)$. Specifically, following previous notations, we have below theorems:
\begin{theorem}[\textsc{Estimation with zero-shot classifier}]\label{theorem:zeroshot}
Let $\mathcal{G}, \mathcal{H}$ be the hypothesis space of vision encoder $g$ and language encoder $h$. Let us denote the rademacher complexity as $\mathfrak{R}(\mathcal{G}\circ\mathcal{H})$ of the combined CLIP model. Supposing the range of $L$ from \cref{eq:clip_loss} as $[0, l^{clip}_{\infty}]$ for all ($\vx,\vz$) in $\mathrm{sup}(Q)$ with $g,h \in \mathcal{G}, \mathcal{H}$. Then, for any $\delta >0$, with probability at least $1-\delta$ we have the following hold:
{\footnotesize
\begin{equation*}
d(P_{zeroshot}, P) \leq \varepsilon_{domain} + \Delta\textcolor{gray}{(\lambda_0\varepsilon_{clip} + \lambda_1 \mathfrak{R}(\mathcal{G}\circ\mathcal{H}) + \lambda_2 l^{clip}_{\infty}\sqrt{\frac{\log 1/\delta}{M}} + \lambda_3\varepsilon_{n})}
\end{equation*}}

\noindent with $\lambda_0, \lambda_1, \lambda_2, \lambda_3 > 0$. Here, $\varepsilon_{domain}$ denotes the bias term induced by the domain gap between $Q$ and $P^{true}$, $\varepsilon_{clip}$ denotes the expected risk of the Bayes optimal CLIP model, and $\Delta \geq 1$ denotes the bias coefficient induced by designing prompts and sampling in \cref{eq:zeroshot}.
\end{theorem}
\begin{theorem}[\textsc{Estimation with induced classifier}]\label{theorem:noisy_classifier}
Let $\mathcal{F}$ be the hypothesis space of induced classifier $f'$. Let us denote the rademacher complexity as $\mathfrak{R}(\mathcal{F})$ of the induced classifier. Supposing the range of $L$ for training $f'$ as $[0, l^{noisy}_{\infty}]$ for all ($\vx,y$) in $\mathrm{sup}(P)$ with $f'\in\mathcal{F}$. Then, for any $\delta >0$, with probability at least $1-\delta$ we have the following holds:
\begin{equation*}
d(P_{induced}, P) \leq \varepsilon_{noise} +  {\color{gray}{\lambda_0 \varepsilon_{induced} + \lambda_1 \mathfrak{R}(\mathcal{F}) + \lambda_2 l^{noisy}_{\infty}\sqrt{\frac{\log 1/\delta}{N}}}}
\end{equation*}
with $\lambda_0, \lambda_1, \lambda_2 > 0$. Here, $\varepsilon_{noise}$ denotes the difference term induced by the label noise in the training dataset, and $\varepsilon_{induced}$ denotes the expected risk of the Bayes optimal induced classifier. 
\end{theorem}
Please refer to \textsc{Supplementary F} for full derivation. With \cref{theorem:zeroshot} and \cref{theorem:noisy_classifier}, ignoring the uncontrollable and common bound error terms (marked in \textcolor{gray}{gray}), we find that \textit{the zero-shot classifier is affected by domain gap and prompts quality while the induced classifier is affected by the label noise of the noisy dataset}, which is intuitively consistent with our expectation\footnote{Please note we are not aiming for strict/tight bounds but to validate the intuition: zero-shot classifier is noise-free while the induced classifier is noise-affected.}.
Put simply, $\varepsilon_{noise}$ is always unavoidable in \cref{theorem:noisy_classifier}, even with a perfectly-learned feature encoder; By contrast, in \cref{theorem:zeroshot}, $\Delta$ can be reduced through better prompt engineering, and $\varepsilon_{domain}$ can be minimized by training CLIP with a more diverse dataset, thus reducing the domain gap. \textit{We emphasize that this is the unique advantage of CLIP for sample selection as a vision language model.}

\subsection{Additional discussion}\label{sec:extra_discussion}

\paragraph{To be greedy or conservative?}
So far, we have mentioned two different conditional probability estimation options~(\cref{eq:estimation_zeroshot} and \cref{eq:estimation_noisy}) and two different sample selection strategies~(\cref{eq:consselection} and \cref{eq:lossselection}), resulting in a total of four different combinations for a possible overall method. The theoretical analysis above and the subsequent empirical ablations show that these different combinations exhibit their preferences in different scenarios. 
In this work, we adopt a conservative strategy by taking the intersection of different sample selection results, prioritizing the precision of sample selection. Compared to more greedy sample selection strategies, we tend to rely on the introduced semi-supervised learning strategy - \textit{MixFix} - to gradually incorporate more samples into training. This can avoid amplifying the impact of noisy samples due to overly greedy sample selection, but it also has the obvious weakness that it will inevitably miss some `hard' clean samples. We leave further exploration to future work.

\paragraph{To fully explore CLIP?}The utilization of the CLIP model for learning with noisy labels remains an area that requires further investigation. To ensure a fair comparison with existing work, we adopt standard sample selection paradigm, refraining from training or fine-tuning the CLIP model~\cite{CoOp,PLOT}. 
The current prominent research directions related to CLIP involve fine-tuning the model, specifically through prompt-based learning. However, as expected, recent work~(CoOp) has indicated that direct fine-tuning CLIP with noisy datasets can yield poorer performance compared to the initial zero-shot classifier. 
Therefore, in addition to sample selection, incorporating established techniques for LNL into prompt-based learning with CLIP may also offer promising directions.

\section{Experiments}
In this section, we conduct extensive experiments on two standard benchmarks with synthetic label noise, CIFAR10 and CIFAR100, and four real-world noisy datasets, Red Mini-ImageNet~\cite{mentormix}, Clothing1M~\cite{clothing1mdataset}, WebVision~\cite{webvision}, and ANIMAL-10N~\cite{selfie}. We mainly follow previous works~\cite{dividemix,instancegm,ssr} for model and training configurations, please refer to \textsc{Supplementary G} for full details. For comparison to other works, we report the results from most advanced SOTA methods - normally including techniques like co-training, contrastive learning, etc.

\begin{figure*}[htbp]
\centering
\resizebox{0.92\textwidth}{!}{%
\begin{minipage}{0.38\textwidth}
    \centering
    \captionof{table}{Ablations on \textit{MixFix} with synthetic CIFAR100 noisy dataset. The \textit{top-3} results are bolded.\label{tab:thetar}}
\resizebox{\textwidth}{!}{%
\begin{tabular}{cccccc}
\toprule
\multirow{2}{*}{$\theta_r$} & \multirow{2}{*}{$\theta'_r$} & \multicolumn{4}{c}{Noise ratio}   \\ \cmidrule(l){3-6} 
                          &                            & 20\%           & 50\%           & 80\%           & 90\%           \\ \midrule
                            & 0.7                         & 76.46          & 74.69          & \textbf{69.50} & 62.91          \\
0.7 & 0.8                         & \textbf{76.63} & \textbf{75.23} & \textbf{69.72} & \textbf{63.11} \\
                            & 0.9                         & \textbf{77.06} & \textbf{75.17} & 67.76          & 59.17          \\ \midrule
                            & 0.7                         & 75.49          & 74.30          & 67.95          & \textbf{63.29} \\
                            & 0.8                         & 76.36          & \textbf{74.90} & 68.86          & \textbf{63.42} \\
\multirow{-3}{*}{0.8}       & 0.9                         & \textbf{76.66} & 74.50          & 67.37          & 58.09          \\ \midrule
                            & 0.7                         & 74.53          & 73.49          & 68.74          & 62.22          \\
                            & 0.8                         & 75.98          & 74.25          & \textbf{68.94} & 62.81          \\
\multirow{-3}{*}{0.9}       & 0.9                         & 75.78          & 74.23          & 67.17          & 59.38          \\ \bottomrule
\end{tabular}%
}

\end{minipage}
\hspace{1em}
\begin{minipage}{0.6\textwidth}
\centering
\includegraphics[width=\linewidth]{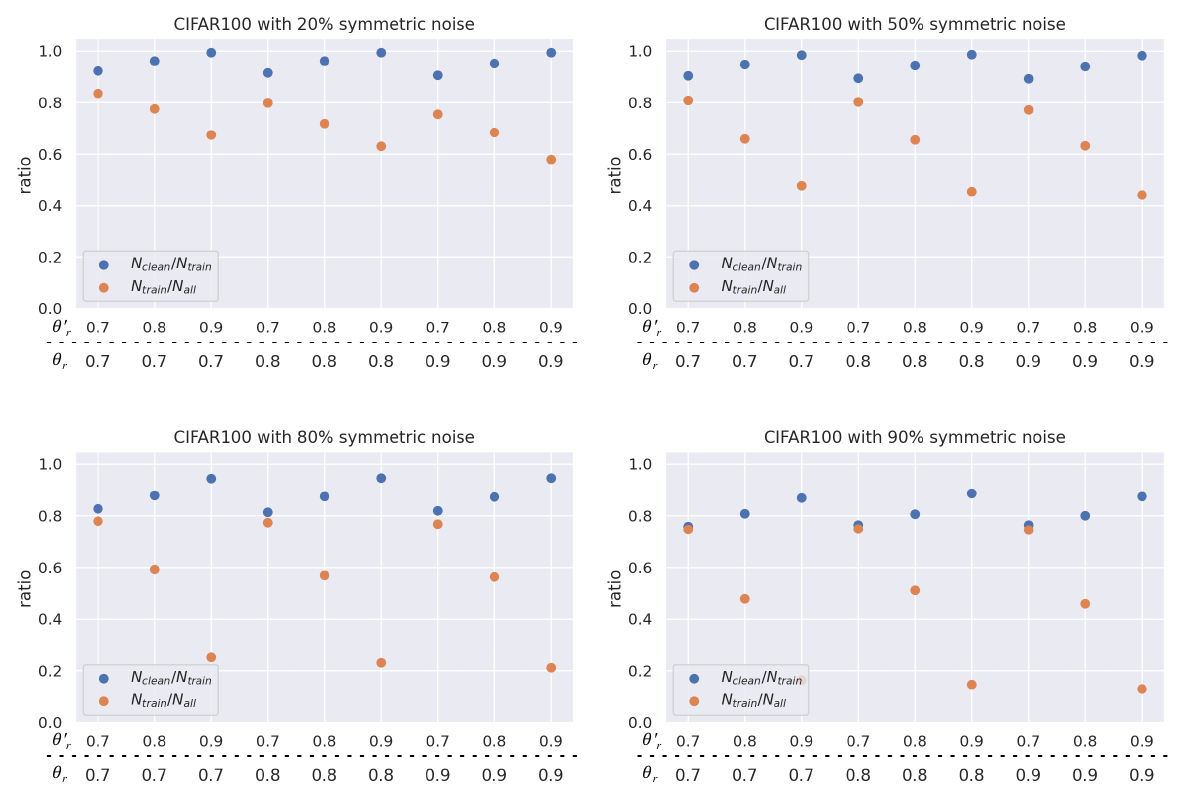}
\captionof{figure}{$N_{train}$ denotes number of training samples, $N_{clean}$ denotes number of clean training samples and $N_{all}$ denotes number of clean training samples. \label{fig:thetar} }
\end{minipage}%
}
\end{figure*}

\begin{table*}[htbp]
\caption{Testing accuracy~(\%) with CLIP zero-shot classifier.}

\centering
\begin{tabular}{lcccccc}
\toprule
Model                       & CIFAR10        & CIFAR100       & Red Mini-ImageNet     & WebVision      & Clothing1M     & ANIMAL-10N     \\ \midrule
CLIP                        & 89.97          & 63.72          & 78.12                 & 73.36          & 39.73          & 76.12          \\ \midrule
SOTA                        & 92.68~\cite{tcl} & 67.7 \cite{tcl}  & 49.55 \cite{instancegm} & 80.9 SSR+ \cite{ssr}  & 74.84 C2D \cite{c2d} & 88.5 SSR+ \cite{ssr}  \\
\rowcolor[HTML]{E4E4E4}Ours & \textbf{95.15} & \textbf{71.17} & \textbf{54.21}        & \textbf{81.56} & \textbf{74.87} & \textbf{88.85} \\ \bottomrule
\end{tabular}%
\label{tab:zeroshot}
\end{table*}

\begin{figure*}[htbp]
  \centering
  \begin{subfigure}{0.33\linewidth}
    \includegraphics[width=\linewidth]{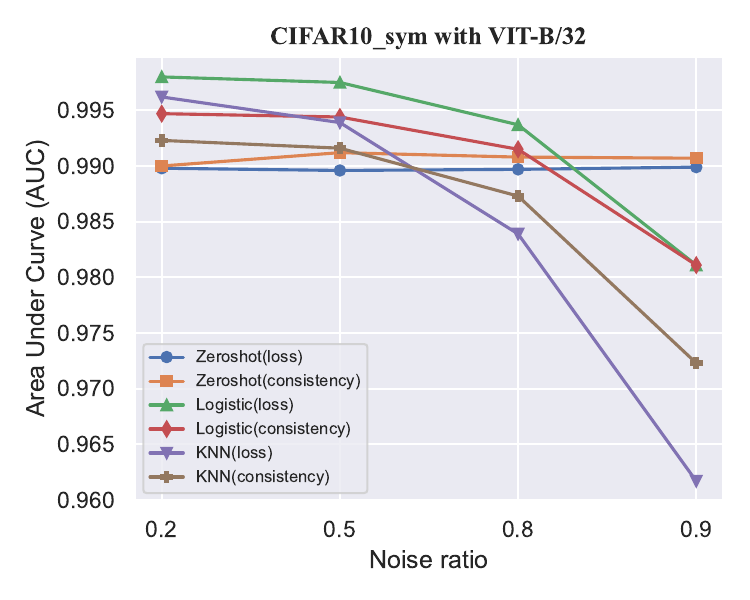}
  \end{subfigure}
    \begin{subfigure}{0.33\linewidth}
    \includegraphics[width=\linewidth]{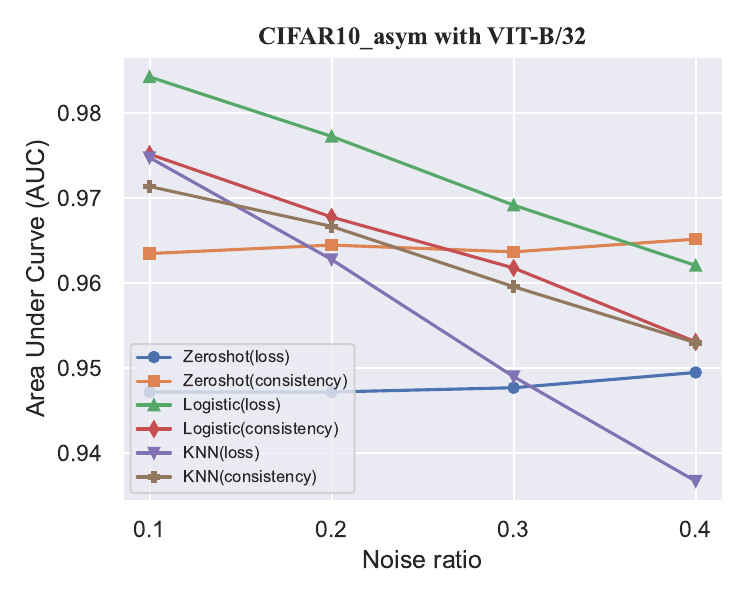}
  \end{subfigure}
  \begin{subfigure}{0.33\linewidth}
    \includegraphics[width=\linewidth]{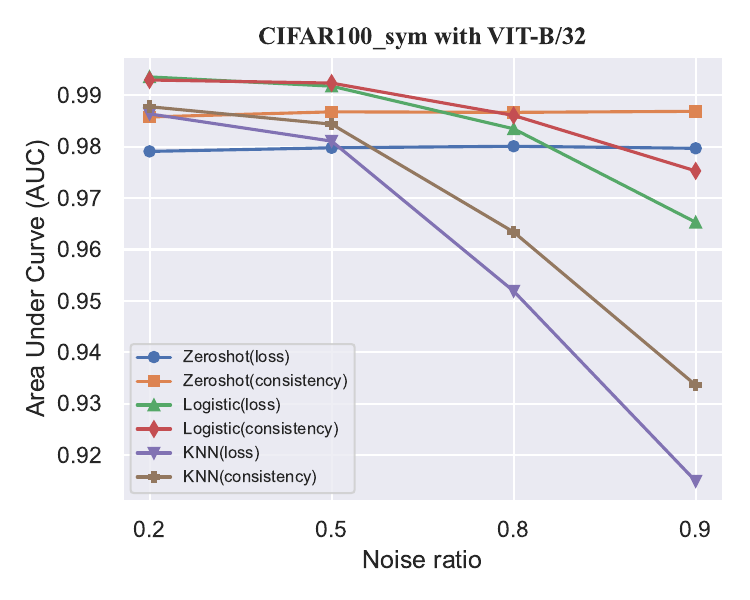}
  \end{subfigure}

  \medskip  

  \begin{subfigure}{0.33\linewidth}
    \includegraphics[width=\linewidth]{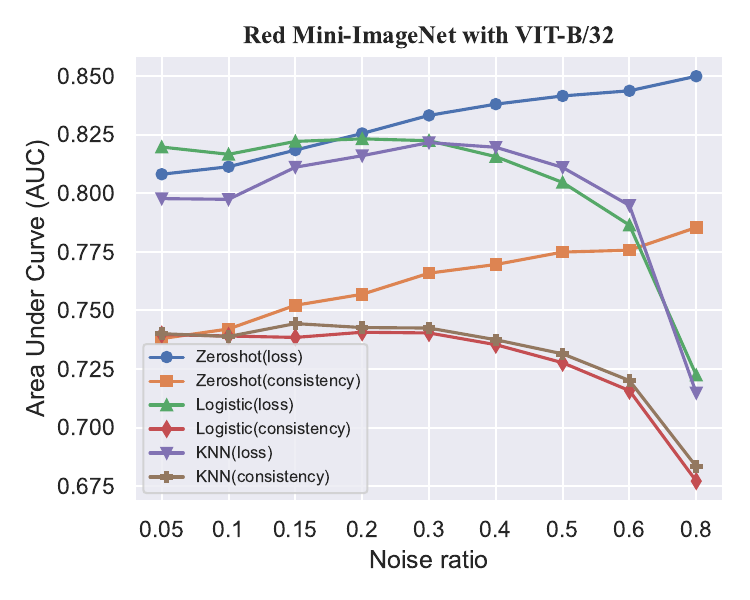}
  \end{subfigure}
    \begin{subfigure}{0.33\linewidth}
    \includegraphics[width=\linewidth]{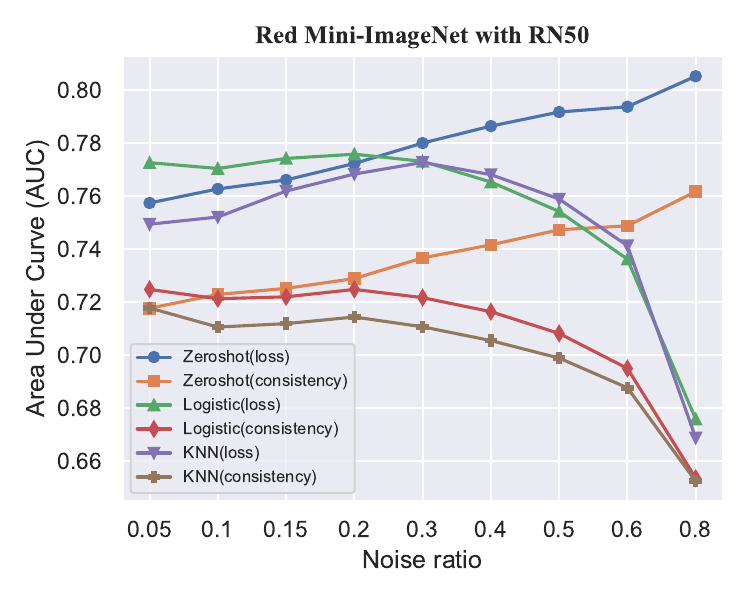}
  \end{subfigure}
  \begin{subfigure}{0.33\linewidth}
    \includegraphics[width=\linewidth]{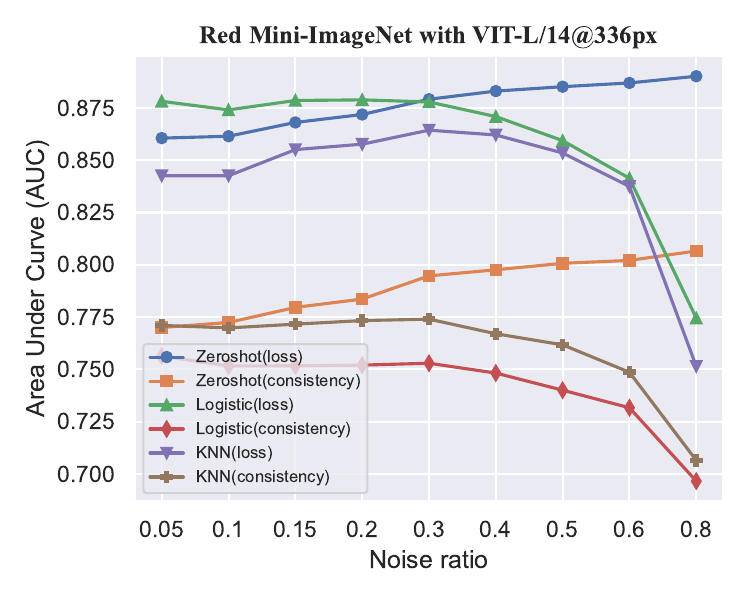}
  \end{subfigure}
    \caption{Comparisons of various sample selection methods \textit{w.r.t} different dataset/noise type/noise ratio. Here, we show the ROC AUC score of binary identification of clean samples.\label{fig:sampleselection_rocauc}}
\end{figure*}

\subsection{Ablations study}\label{sec:exp_ablation}
\paragraph{Hyper-parameters \textit{w.r.t} \textit{MixFix}}
\label{appdix:semi-sup_hyper}
In this section, we ablate on the only two hyperparameters of our semi-supervised training strategy \textit{MixFix}: the `absorb' threshold $\theta_r$ and the `relabel' threshold $\theta'_r$. Owing to the precision-recall dilemma when doing sample selection, here we also need to weigh the precision and recall when introducing additional training samples. 
In \cref{tab:thetar} we demonstrate that under different noise ratios, a too-high or too-low threshold leads to performance degradation, and 
$\theta_r < \theta_r'$ leads to better performance than setting the same value for both thresholds. In \cref{fig:thetar}, we further reveal the inherent mechanism. Especially, after reducing the `\texttt{absorb}' threshold $\theta'_r$, the proportion of training samples increases and the accuracy of training samples decreases.

\paragraph{Analyzing CLIP Zero-shot classification as a baseline}
\label{exp: 2_clip_baseline}
In this section, we consider utilizing CLIP's zero-shot classifier directly on the clean test set, following a procedure that we describe in \Cref{sec:meth_robustclip}. In \cref{tab:zeroshot}, we present the zero-shot classification results on six involved benchmarks and compare them with current SOTA results as well as our own method. It's worth noting that CLIP is utilized with the VIT-B/32 architecture here, while our method and the SOTA methods adopt simpler structures, such as PreResNet-18 for the CIFAR dataset. Therefore, this comparison is indeed `over stringent'. Even though, we observe that, when compared to directly utilizing CLIP's zero-shot classifier, our method delivers significant improvements on most datasets and outperforms the SOTA LNL methods on all datasets.
We also consider other vision-language models other than CLIP in \textsc{Supplementary A}.

\paragraph{Analyzing sample selection \textit{w.r.t} different classifiers and different mechanisms}
In \cref{sec:theory}, we theoretically conclude that the sample selection performance of the zero-shot classifier is influenced by the quality of utilized prompts and the domain gap between CLIP training dataset and the in-question noisy dataset, while the performance of the easily-induced classifier trained based on CLIP's vision encoder and the in-question noisy dataset is influenced by the noise of the in-question dataset. To validate this, we empirically test with two datasets with controllable noise ratios, that is, the CIFAR10/100 dataset with synthetic noise and the Red Mini-ImageNet dataset with real-world noise.

In \cref{fig:sampleselection_rocauc}, we show the sample selection result and find that:
\begin{itemize}
    \item Firstly, as the noise ratio increases, regardless of the dataset (CIFAR10 \textit{vs.} CIFAR100 \textit{vs.} Red Mini-ImageNet), noise modes (symmetric \textit{vs.} asymmetric \textit{vs.} real-world) or CLIP backbones (VIT-B/32 \textit{vs.} VIT-L/14@336px \textit{vs.} RN50), the zero-shot classifier gradually outperforms the induced classifier. This further validates the unique advantage of CLIP and our theoretical findings in \cref{sec:theory} - that the latter is affected by label noise while the former is not;
    \item Additionally, we find that different sample selection mechanisms ($\mathbb{G}_{consistency}$ VS $\mathbb{G}_{loss}$) show distinct advantages and disadvantages on different datasets. Given that noise information is typically unknown in real-world scenarios, as analyzed in \cref{sec:extra_discussion}, we default to a conservative sample selection strategy, which involves utilizing both sample selection strategies and choosing their intersection as final selected subset;
    \item Furthermore, we notice that when comparing two different choices for obtaining the induced classifier, the \textit{LogisticRegression} classifier empirically exhibits superior performance to the \textit{kNN} classifier. Therefore, we choose the \textit{LogisticRegression} classifier as our default choice for the induced classifier.
\end{itemize}

\subsection{Results on synthetic noisy dataset}
In this section, we first evaluate our method on the CIFAR datasets with synthetic symmetric/asymmetric noise. In \cref{tab:cifar}, We can see that our method gets competitive and better performance in all experiment settings, especially when the noise ratio is high~(63.11\% testing accuracy with 90\% symmetric noise on CIFAR100 dataset).
Also, we would like to emphasize that we keep hyper-parameters fixed for all experiments here as we believe the method robustness in a noise-agnostic scenario is critical. 

To further validate the performance of our method in handling the `hard noise', we also conduct experiments on instance-dependent noise in~\cref{tab:instancenoise}. Different from symmetric or asymmetric noise, instance-dependent noise assumes that semantic-similar samples are more prone to get mislabelled, aligning better with our earlier definition of `hard noise'. 
Besides, here we here exclude \textit{MixFix} and employ the selected samples for training with cross-entropy loss solely. This exclusion serves to provide additional proof of the superior sample selection performance of \textit{CLIPCleaner}.

\begin{table}[htbp]
\caption{Testing accuracy (\%) on CIFAR10 with instance-dependent noise. \label{tab:instancenoise}}
\resizebox{\linewidth}{!}{%
\begin{tabular}{lcccc}
\toprule
\multirow{2}{*}{Method} & \multicolumn{4}{c}{Noise ratio}                                                                                                                   \\ \cmidrule(l){2-5} 
                        & 10\%                                & 20\%                                & 30\%                                & 40\%                                \\ \midrule
CE                      & 91.25                              & 86.34                              & 80.87                              & 75.68                              \\
F-correction~\cite{loss_correction}                 & 91.06                              & 86.35                              & 78.87                              & 71.12                              \\
Co-teaching~\cite{coteaching}             & 91.22                              & 87.28                              & 84.33                              & 78.72                              \\
GCE~\cite{generalized_cross_entropy}                     & 90.97                              & 86.44                              & 81.54                              & 76.71                              \\
DAC~\cite{thulasidasan2019DAC}                     & 90.94                              & 86.16                              & 80.88                              & 74.80                              \\
DMI~\cite{xu2019l_dmi}                     & 91.26                              & 86.57                              & 81.98                              & 77.81                              \\
SEAL~\cite{chen2021INDnoise}                    & 91.32                              & 87.79                              & 85.30                              & 82.98                              \\ \midrule
CE*                     & 90.76          & 86.08          & 80.64          & 75.27         \\ 
\rowcolor[HTML]{E4E4E4} CLIPCleaner + CE                    & \textbf{92.33$\pm$0.37} & \textbf{91.06$\pm$0.37} & \textbf{89.71$\pm$0.37} & \textbf{88.26$\pm$0.37} \\ \bottomrule
\end{tabular}%
}

\end{table}

\begin{table*}[htbp]
\caption{Testing accuracy~(\%) on CIFAR-10 and CIFAR-100 with synthetic noise.\label{tab:cifar}} 
\begin{center}
\resizebox{1\textwidth}{!}{
\begin{tabular}{lccccccccc}
\toprule
Dataset                                & \multicolumn{5}{c}{CIFAR10}                                                                                      & \multicolumn{4}{c}{CIFAR100}                                                              \\ \cmidrule(l){2-6} \cmidrule(l){7-10}
Noise type                             & \multicolumn{4}{c}{Symmetric}                                                             & Assymetric           & \multicolumn{4}{c}{Symmetric}                                                             \\ \cmidrule(l){2-5} \cmidrule(l){6-6} \cmidrule(l){7-10}
Noise ratio                            & 20\%                 & 50\%                 & 80\%                 & 90\%                 & 40\%                 & 20\%                 & 50\%                 & 80\%                 & 90\%                 \\ \midrule
CE                          & 86.8                 & 79.4                 & 62.9                 & 42.7                 & 85.0                 & 62.0                 & 46.7                 & 19.9                 & 10.1                 \\
Co-teaching+~\cite{coteaching+}       & 89.5                 & 85.7                 & 67.4                 & 47.9                 & -                    & 65.6                 & 51.8                 & 27.9                 & 13.7                 \\
F-correction~\cite{loss_correction}   & 86.8                 & 79.8                 & 63.3                 & 42.9                 & 87.2                 & 61.5                 & 46.6                 & 19.9                 & 10.2                 \\
PENCIL~\cite{pencil}                  & 92.4                 & 89.1                 & 77.5                 & 58.9                 & 88.5                 & 69.4                 & 57.5                 & 31.1                 & 15.3                 \\
LossModelling~\cite{lossmodellingbmm} & 94.0                 & 92.0                 & 86.8                 & 69.1                 & 87.4                 & 73.9                 & 66.1                 & 48.2                 & 24.3                 \\
DivideMix~\cite{dividemix}            & \textbf{96.1}  & 94.6                    & 93.2                    & 76.0                    & 93.4                    & 77.3           & 74.6           & 60.2                    & 31.5                    \\
ELR+~\cite{elr}                       & 95.8           & 94.8                    & 93.3                    & 78.7                    & 93.0                    & 77.6           & 73.6           & 60.8                    & 33.4                    \\
MOIT~\cite{moit}                      & 93.1           & 90.0                    & 79.0                    & 69.6                    & 92.0                    & 73.0           & 64.6           & 46.5                    & 36.0                    \\
SelCL+~\cite{selcl}                   & 95.5           & 93.9                    & 89.2                    & 81.9                    & 93.4                    & 76.5           & 72.4           & 59.6                    & 48.8                    \\
TCL~\cite{tcl}                        & 95.0           & 93.9                    & 92.5                    & 89.4                    & 92.6                    & 78.0           & 73.3           & 65.0                    & 54.5                    \\ \midrule
\rowcolor[HTML]{E4E4E4} Ours            & 95.92$\pm$0.15 & \textbf{95.67$\pm$0.28} & \textbf{95.04$\pm$0.37} & \textbf{94.23$\pm$0.54} & \textbf{94.89$\pm$0.16} & \textbf{78.20$\pm$0.45} & \textbf{75.23$\pm$0.29} & \textbf{69.72$\pm$0.61} & \textbf{63.11$\pm$0.89} \\ \bottomrule
\end{tabular}%
}

\end{center}
\end{table*}

\begin{table*}[htbp]
\caption{Testing accuracy~(\%) on Clothing1M.\label{tab:clothing}}
\centering
\resizebox{\linewidth}{!}{%
\begin{tabular}{cccccccc
>{\columncolor[HTML]{E4E4E4}}c 
>{\columncolor[HTML]{E4E4E4}}c 
>{\columncolor[HTML]{E4E4E4}}c}
\toprule
CE & \begin{tabular}[c]{@{}c@{}}F-correction\\ \cite{loss_correction}\end{tabular} & \begin{tabular}[c]{@{}c@{}}RRL\\ \cite{rrl}\end{tabular} & \begin{tabular}[c]{@{}c@{}}C2D\\ \cite{c2d}\end{tabular} & \begin{tabular}[c]{@{}c@{}}DivideMix\\ \cite{dividemix}\end{tabular} & \begin{tabular}[c]{@{}c@{}}ELR+\\ \cite{elr}\end{tabular} & \begin{tabular}[c]{@{}c@{}}SSR+\\ \cite{ssr}\end{tabular} & \begin{tabular}[c]{@{}c@{}}TCL\\ \cite{tcl}\end{tabular} & Ours           & Ours~(Co-training) & CLIPCleaner + DivideMix    \\ \midrule
69.21         & 69.84                                                                         & 74.30                                                    & 74.84                                                    & 74.76                                                                & 74.81                                                     & 74.83                                                     & 74.80                                                    & 73.41$\pm$0.65 & 74.01$\pm$0.47     & \textbf{74.87$\pm$0.44} \\ \bottomrule
\end{tabular}%
}
\end{table*}

\subsection{Results on real-world noisy datasets}
Finally, in~\cref{tab:webvision},~\cref{tab:redimagenet}, and~\cref{tab:animal10n} we show results on the ANIMAL-10N, Red Mini-ImageNet and WebVision datasets, respectively. In summary, our proposed method demonstrates substantial improvements compared to the current state-of-the-art approaches on both large-scale web-crawled datasets and small-scale human-annotated noisy datasets. 

\begin{table}[htbp]
\caption{Testing accuracy~(\%) on WebVision.\label{tab:webvision}}

\begin{center}
\centering
\resizebox{1.0\linewidth}{!}{%
\begin{tabular}{lcccc}
\toprule
\multirow{2}{*}{Methods}       & \multicolumn{2}{c}{WebVision} & \multicolumn{2}{c}{ILSVRC2012} \\ \cmidrule(l){2-3}\cmidrule(l){4-5}  
                               & Top1          & Top5          & Top1           & Top5          \\ \midrule
Co-teaching~\cite{coteaching} & 63.5          & 85.20         & 61.48          & 84.70         \\
DivideMix~\cite{dividemix}    & 77.32         & 91.64         & 75.20          & 90.84         \\
ELR+~\cite{elr}               & 77.78         & 91.68         & 70.29          & 89.76         \\
NGC~\cite{ngc}                & 79.16         & 91.84         & 74.44          & 91.04         \\
FaMUS~\cite{famus}                          & 79.4          & 92.8          & 77.0           & 92.8          \\
RRL~\cite{rrl}                & 76.3          & 91.5          & 73.3           & 91.2          \\
SelCL+~\cite{selcl}                         & 79.9          & 92.6          & 76.8           & \textbf{93.0}          \\
SSR+~\cite{ssr}                           & 80.9          & 92.8          & 75.8           & 91.8          \\
TCL~\cite{tcl}                            & 79.1          & 92.3          & 75.4           & 92.4          \\ \midrule
\rowcolor[HTML]{E4E4E4} 
Ours  &    \textbf{81.56$\pm$0.29}           &      \textbf{93.26$\pm$0.65}         &   \textbf{77.80$\pm$0.25}             &      92.08$\pm$0.44         \\ \bottomrule
\end{tabular}}

\end{center}
\end{table}

\begin{table}[htbp]
\caption{Testing accuracy~(\%) on Red Mini-ImageNet.\label{tab:redimagenet}}

\centering
\resizebox{1.0\linewidth}{!}{%
\begin{tabular}{lcccc}
\toprule
\multirow{2}{*}{Method} & \multicolumn{4}{c}{Noise ratio} \\ \cmidrule(l){2-5} 
                        & 20\%   & 40\%   & 60\%  & 80\%  \\ \midrule
CE            & 47.36  & 42.70  & 37.30 & 29.76 \\
Mixup~\cite{Mixup}                   & 49.10  & 46.40  & 40.58 & 33.58 \\
DivideMix~\cite{dividemix}               & 50.96  & 46.72  & 43.14 & 34.50 \\
MentorMix~\cite{mentormix}               & 51.02  & 47.14  & 43.80 & 33.46 \\
FaMUS~\cite{famus}                   & 51.42  & 48.06  & 45.10 & 35.50 \\
InstanceGM~\cite{instancegm}              & 58.38  & 52.24  & 47.96 & 39.62 \\ \midrule
\rowcolor[HTML]{E4E4E4} 
Ours                    &   \textbf{61.44$\pm$0.45}     &  \textbf{58.42$\pm$0.66}     & \textbf{53.18$\pm$0.47} &   \textbf{43.82$\pm$0.87}    \\ \bottomrule
\end{tabular}
}
\end{table}

\begin{table}[htbp]
\caption{Testing accuracy~(\%) on ANIMAL-10N. \label{tab:animal10n}} 
\centering
\resizebox{0.45\linewidth}{!}{%
\begin{tabular}{l c}
\toprule
Method & Accuracy \\
\midrule
CE & 79.4 \\
SELFIE~\cite{selfie} & 81.8 \\
PLC~\cite{PLC} & 83.4 \\
NCT~\cite{nestedcoteaching} & 84.1 \\
InstanceGM~\cite{instancegm} & 84.6 \\ 
SSR+~\cite{ssr} & 88.5 \\ \midrule
\rowcolor[HTML]{E4E4E4} Ours & \textbf{88.85$\pm$0.61} \\
\bottomrule
\end{tabular}}
\end{table}

We note, that the proposed \textit{CLIPCleaner} can also be used in combination with other schemes. In \cref{tab:clothing} we show results on the Clothing1M dataset both with our default setting~(\textit{CLIPCleaner + MixFix}) and with it incorporated to two additional schemes: first incorporating our method with co-training, and second replacing \textit{MixFix} with DivideMix~\cite{dividemix}. We observe that we obtain results that are superior to the current state-of-the-art. Meanwhile, we would like to note that the majority of existing methods have small differences on the Clothing1M dataset despite the fact that they have large performance differences on other datasets. This suggests that additional training techniques may have a greater impact than sample selection methods on this specific dataset, possibly due to the fact that the Clothing1M dataset is more fine-grained than other datasets.  For such fine-grained noisy datasets, sample selection may not be the optimal strategy, as suggested in \textsc{Supplementary H}.

\section{Conclusion}
\label{sec:conclusion}
To mitigate the issues of `self-confirmation bias' and compensate for visual-only modality in current mainstream sample selection methods, in this paper, we propose a method utilizing the large-scale vision-language model CLIP for sample selection, called \textit{CLIPCleaner}. We substantiate its effectiveness both theoretically and empirically. Furthermore, we introduce a straightforward semi-supervised learning method tailored for noisy datasets, called \textit{MixFix}, without the need for intricate off-the-shelf techniques. We emphasize that the exploration of utilizing vision-language models for noisy datasets, such as the potential of existing prompt learning techniques, remains an open direction. Additionally, the possibility of a large domain gap between the CLIP model and the target dataset can influence results, indicating a need for more refined vision-language models. Lastly, our experiments suggest that sample selection methods may not be optimal for fine-grained noisy datasets, which presents itself also as one of our future research directions.

\bibliographystyle{ACM-Reference-Format}

\appendix
\def\thesection{\Alph{section}}
\section{Sample selection with other vision-language models}
\label{appdix:other}
Here, we compare CLIP with another vision-language model - ALIGN~\cite{jia2021}. Specifically, we compare their performance on sample selection based on the CIFAR10 dataset with instance-dependent noise~\cite{chen2021INDnoise}. In \cref{tab:alignvsclip}, we can see ALIGN behaves similarly well as CLIP concerning precision with even higher recall. This demonstrates that our proposed idea of using vision-language models for sample selection is widely effective.

\begin{table}[htbp]
\caption{Precision-Recall of sample selection results on CIFAR10 with instance-dependent noise with CLIP and ALIGN.\label{tab:alignvsclip}}
\centering
\resizebox{\linewidth}{!}{%
\begin{tabular}{lcccccccc}
\toprule
Noise ratio & \multicolumn{2}{c}{0.1} & \multicolumn{2}{c}{0.2} & \multicolumn{2}{c}{0.3} & \multicolumn{2}{c}{0.4} \\ \midrule
            & Precision    & Recall   & Precision    & Recall   & Precision    & Recall   & Precision    & Recall   \\
CLIP        & 99.73        & 70.75    & 99.53        & 75.07    & 99.25        & 77.77    & 99.03        & 79.23    \\
ALIGN       & 99.47        & 72.47    & 99.13        & 78.64    & 99.01        & 81.22    & 98.74        & 84.53    \\ \bottomrule
\end{tabular}%
}
\end{table}

\section{Prompts generation and further analysis}
\label{appdix:prompts}

\subsection{How multiple prompts with class-specific features are generated?}
Regarding the generation multiple prompts based on class-specific features, motivated by recent work~\cite{visualdescription}, we first generate multiple features for each class by asking ChatGPT about each category's characteristics. We use below question for ChatGPT 3.5:
\begin{center}
    \texttt{
For CLIP model, the prompts matter a lot. can you give me some discriminative features of some classes? Please list it in nested python as each class has multiple descriptions. Please ensure it is formatted as `which has ...' or `which is ...' or `which ...'. For example, [`Cat', `Lynx', `Wolf', `Coyote', `jaguar', `Cheetah', `Chimpanzee', `Orangutan', `Hamster', `Guinea pig'].
}
\end{center}

 We then generate multiple prompts with template in \textsc{Section 3.2}. We will include our generated class-specific prompts along with the code upon acceptance.

\subsection{Comparison of class-specific prompts with other prompt style}
To experimentally validate the superiority of our prompt style based on class-specific features, we conduct a comparative analysis of its zero-shot classification performance against alternative prompt styles. Specifically, we consider three empirical variants including ours:
\begin{enumerate}
    \item Single prompt: `\texttt{A photo of \{class name of $y_i$\}.}';
    \item Multiple prompts with different templates: `\texttt{A good photo of \{class name of $y_i$\}.}'/`\texttt{An old picture of \{class name of $y_i$\}.}' .etc;
    \item Multiple prompts with class-specific features: `\texttt{A photo of \{class name of $y_i$\}, which is/has \{class-specific feature $j$ of class $y_i$\}.}' with features such as the color, shape, etc.
\end{enumerate}

In \cref{tab:ablation_zeroshot}, we present zero-shot classification results on six noisy datasets using the three prompt styles mentioned above and different backbones for CLIP model~(VIT-B/32 and VIT-L/14@336px). We observe that, in most cases, the effectiveness of our prompting style is at its best, especially when employing a larger-scale CLIP backbone~(VIT-L/14@336px). This aligns with our theoretical analysis.

\begin{table}[htbp]
\caption{Zero-shot classification with different prompt styles.\label{tab:ablation_zeroshot}}
\resizebox{\linewidth}{!}{%
\begin{tabular}{lccccccc}
\toprule
Model                                            & Prompt technique & CIFAR10        & CIFAR100       & Red Mini-ImageNet & WebVision      & Clothing1M     & ANIMAL-10N     \\ \midrule
\multirow{3}{*}{CLIP (ViT-B/32)}       & 1                & 88.29          & 61.62          & 74.40             & 72.40          & 39.80          & 75.08          \\
                                                 & 2                & 89.73          & 63.65          & 75.14             & 68.12          & 39.68          & 75.70          \\
                                                 & 3                & 87.97          & 63.72          & 78.12             & 73.36          & 37.73          & 74.62          \\ \midrule
\multirow{3}{*}{CLIP (ViT-L/14@336px)} & 1                & 94.78          & 74.36          & 80.20             & 45.13          & 85.18          & 85.12          \\
                                                 & 2                & 95.17          & 74.96          & 79.88             & 47.26          & \textbf{85.78} & 87.00          \\
                                                 & 3                & \textbf{95.19} & \textbf{76.78} & \textbf{81.96}    & \textbf{48.15} & 85.36          & \textbf{87.98} \\ \bottomrule
\end{tabular}%
}
\end{table}

\section{Utilizing \textit{CLIPCleaner} with other methods}
Current mainstream methods for learning with noisy labels usually involve an iterative process consists of sample selection and model training. Normally, these methods require a warm-up stage, i.e., training with whole dataset for some epochs to learn an usable model before the iteration process. Here, we consider to utilize the selected samples by \textit{CLIPCleaner} for the warmup stage to validate if it can also bring improvement for existing methods. In \cref{tab:dmixcc}, we validate that \textit{CLIPCleaner} brings steady improvement over original DivideMix~\cite{dividemix}.

\begin{table*}[htbp]
\caption{Testing accuracy~(\%) of \textit{CLIPCleaner} utilized along with DivideMix.\label{tab:dmixcc}} 
\begin{center}
\resizebox{1\textwidth}{!}{
\begin{tabular}{lccccccccc}
\toprule
Dataset                                & \multicolumn{5}{c}{CIFAR10}                                                                                      & \multicolumn{4}{c}{CIFAR100}                                                              \\ \cmidrule(l){2-6} \cmidrule(l){7-10}
Noise type                             & \multicolumn{4}{c}{Symmetric}                                                             & Assymetric           & \multicolumn{4}{c}{Symmetric}                                                             \\ \cmidrule(l){2-5} \cmidrule(l){6-6} \cmidrule(l){7-10}
Noise ratio                            & 20\%                 & 50\%                 & 80\%                 & 90\%                 & 40\%                 & 20\%                 & 50\%                 & 80\%                 & 90\%                 \\ \midrule
Cross-Entropy                         & 86.8  & 79.4  & 62.9  & 42.7  & 85.0       & 62.0   & 46.7  & 19.9  & 10.1 \\
Co-teaching+~\cite{coteaching+}       & 89.5  & 85.7  & 67.4  & 47.9  & -          & 65.6   & 51.8  & 27.9  & 13.7 \\
F-correction~\cite{loss_correction}   & 86.8  & 79.8  & 63.3  & 42.9  & 87.2       & 61.5   & 46.6  & 19.9  & 10.2 \\
PENCIL~\cite{pencil}                  & 92.4  & 89.1  & 77.5  & 58.9  & 88.5       & 69.4   & 57.5  & 31.1  & 15.3 \\
LossModelling~\cite{lossmodellingbmm} & 94.0  & 92.0  & 86.8  & 69.1  & 87.4       & 73.9   & 66.1  & 48.2  & 24.3 \\
\rowcolor[HTML]{E4E4E4}
DivideMix~\cite{dividemix}            & 96.1  & 94.6  & 93.2  & 76.0  & 93.4       & 77.3   & 74.6  & 60.2  & 31.5 \\
\rowcolor[HTML]{E4E4E4}
DivideMix + CLIPCleaner                  & 96.3~\footnotesize{(\textcolor{ForestGreen}{0.2$\uparrow$})}
  & 95.7~\footnotesize{(\textcolor{ForestGreen}{1.1$\uparrow$})}
  & 94.3~\footnotesize{(\textcolor{ForestGreen}{1.1$\uparrow$})}
  & 88.7~\footnotesize{(\textcolor{ForestGreen}{12.7$\uparrow$})}
  & 94.2~\footnotesize{(\textcolor{ForestGreen}{0.8$\uparrow$})}
       & 78.1~\footnotesize{(\textcolor{ForestGreen}{0.8$\uparrow$})}
   & 75.2~\footnotesize{(\textcolor{ForestGreen}{0.6$\uparrow$})}
  & 71.3~\footnotesize{(\textcolor{ForestGreen}{9.1$\uparrow$})}
  & 46.6~\footnotesize{(\textcolor{ForestGreen}{15.1$\uparrow$})}
 \\ \bottomrule
\end{tabular}%
}

\end{center}
\end{table*}

\section{Per-class seperate GMM vs Whole single GMM}
\label{appdix:singlegmm}
In this section, we compare the differences between using seperate GMM for each class and a single GMM for all classes in sample selection. We conduct experiments on the CIFAR10 dataset with instance-dependent noise. As shown in \cref{tab:pergmm_prrecall}, we observe that the seperate GMM yields a higher recall while maintaining competitive precision in sample selection.
In \cref{tab:pergmm_minmax}, we find and validate that the seperate GMM allows us to obtain a more balanced subset, thereby mitigating class imbalance issues and partially explaining why we achieve a better recall above.
\begin{table}[htbp]
\caption{Precision and recall of sample selection on CIFAR10 dataset with instance-dependent noise with Separate and Single GMM.\label{tab:pergmm_prrecall}}
\centering
\resizebox{\linewidth}{!}{%
\begin{tabular}{lcccccccc}
\toprule
Noise ratio  & \multicolumn{2}{c}{0.1} & \multicolumn{2}{c}{0.2} & \multicolumn{2}{c}{0.3} & \multicolumn{2}{c}{0.4} \\ \midrule
             & Precision    & Recall   & Precision    & Recall   & Precision    & Recall   & Precision    & Recall   \\
Separate GMM & 99.73        & 70.75    & 99.53        & 75.09    & 99.25        & 77.77    & 99.04        & 79.26    \\
Single GMM   & 99.77        & 68.90    & 99.61        & 71.88    & 99.43        & 73.68    & 99.29        & 72.67    \\ \bottomrule
\end{tabular}%
}

\end{table}

\begin{table}[htbp]
\caption{Max-Min number of selected samples from each class.\label{tab:pergmm_minmax}}
\centering
\resizebox{\linewidth}{!}{%
\begin{tabular}{lccccccccc}
\toprule
Noise ratio  & \multicolumn{2}{c}{0.1} & \multicolumn{2}{c}{0.2} & \multicolumn{2}{c}{0.3} & \multicolumn{2}{c}{0.4} \\ \midrule
             & Max        & Min        & Max        & Min        & Max        & Min        & Max        & Min        \\ \midrule
Separate GMM & 4061       & 2228       & 3938       & 2148       & 3656       & 1851       & 3312       & 1440       \\
Single GMM   & 4188       & 1720       & 4038       & 1757       & 3682       & 1455       & 3403       & 947        \\ \bottomrule
\end{tabular}%
}

\end{table}

\section{Relation between joint probability and the CLIP similarity}\label{sec:relation}
In this part, we briefly explain the probabilistic relation of the learned similarity value~($\exp(g(\vx_i)^T h(\vz_i))$) and the joint probability $Q(\rvz=\vz_i,\rvx=\vx_i)$. Specifically, we can easily write the empirical risk with CLIP loss function in eq. (2) as:
{\footnotesize
\begin{align*}
    \hat{R}^{Q}(g, h) &= \frac{1}{2} \sum_{i=1}^M(-\log \frac{\exp(g(\vx_i)^T h(\vz_i))}{\sum_{j=1}^M \exp(g(\vx_j)^T h(\vz_i))} -\log \frac{\exp(g(\vx_i)^T h(\vz_i))}{\sum_{j=1}^M \exp(g(\vx_i)^T h(\vz_j))}) \\
    & = -\frac{1}{2}\log \prod_{i=1}^M \frac{\exp(g(\vx_i)^T h(\vz_i))}{\sum_{j=1}^M \exp(g(\vx_j)^T h(\vz_i))} \frac{\exp(g(\vx_i)^T h(\vz_i))}{\sum_{j=1}^M \exp(g(\vx_i)^T h(\vz_j))}
\end{align*}
}

\noindent For the specific \textit{i.i.d} sampled dataset, based on \textit{MLE principle} we have the negative log-likelihood as:
{\scriptsize
\begin{align*}
    \mathcal{L}(g,h;{(\vx_i,\vz_i)}_{i=1}^{M}) &= -\log \prod_{i=1}^M Q(\rvx = \vx_i| \rvz= \vz_i, \rvx\in \{\vx_j\}_{j=1}^M) Q(\rvz=\vz_i| \vx_i, \rvz\in \{\vz_j\}_{j=1}^M) \\
    &= -\log \prod_{i=1}^M \frac{Q(\rvx = \vx_i, \rvz= \vz_i)}{\sum_{j=1}^M Q(\rvx=\vx_j, \rvz=\vz_i)} \frac{Q(\rvx=\vx_i, \rvz=\vz_i)}{\sum_{j=1}^M Q(\rvx=\vx_i, \rvz=\vz_j)}
\end{align*}
}
Comparing $\hat{R}^{Q}(g, h)$ with $\mathcal{L}(g,h;{(\vx_i,\vz_i)}_{i=1}^{M})$, we have: $$\exp(g(\vx_i)^T h(\vz_j)) \propto Q(\rvx=\vx_i, \rvz=\vz_j),$$ where latter serves as an estimation of $Q(\vx_i, \vz_j)$ after training.

\section{Derivation of \textsc{Theorem 3.1} and \textsc{Theorem 3.2}}
\label{appdix:theory_proof}
In this section, we provide full derivation of \textsc{Theorem 3.1} and \textsc{Theorem 3.2}. 
To start with, we first state the essential generalization error bound based on Rademacher complexity~($\mathfrak{R}$):
\begin{lemma}[Rademacher generalization error bound~\cite{mohri2018foundationmodel}]\label{theorem: rademacher}
    Supposing we have $N$ i.i.d samples $\{\vx_i\}_{i=1}^N$ from distribution $P(\vx)$. Let $\mathcal{F}$ be the hypothesis space of model $f$ and $L$ be any classification-calibrated surrogate loss function of 0-1 loss ranging from $[a, b]$. Then, for any $\delta >0$, with probability at least $1-\delta$  we have the following holds for all $f \in \mathcal{F}$:
    \begin{equation*}
            R(f) \leq \hat{R}(f) + 2\mathfrak{R}(\mathcal{\mathcal{F}}) + (b-a)\sqrt{\frac{\log(1/\delta)}{2N}}
    \end{equation*}
    Here, $R^P(f) = E_{P(\vx)} L(\vx; f)$ denotes the expected risk with $f$ and $\hat{R}^P(f) = \frac{1}{N}\sum_{i=1}^N L(\vx_i; f)$ denotes the empirical one. Please do not confuse the notations here with other notations.
\end{lemma}

\subsection{Derivation of \textsc{Theorem 3.1}}
Let us recall the formulation of the CLIP model. CLIP aims to learn from a dataset of image-text pairs, denoted as ${(\vx_i,\vz_i)}_{i=1}^{M}$, which is \textit{i.i.d.} sampled from a hidden joint distribution $Q(\rvx, \rvz)$ with $\mathrm{sup}(Q) = \{\rvx \in \mathbb{R}^{C\times H\times W}, \rvz\in \mathbb{R}^d\}$. \textit{As the dataset for training CLIP is often also considered `noisy'\footnote{The image description sometimes can be random due to the data collection process~\cite{jia2021,clip}. We here also consider this into consideration. \textbf{Please note this is different from our interested label noise in this work}.}, we denote the clean joint distribution for the CLIP training dataset as $Q'(\rvx, \rvz)$ and the corresponding clean dataset as ${(\vx_i,\vz'_i)}_{i=1}^{M}$.}

To measure the distance between $P_{zeroshot}(\ry|\rvx)$ with $P(\ry|\rvx)$, we divide it into two parts, i.e, the distance between $P_{zeroshot}(\ry|\rvx)$ and $Q'(\ry|\rvx)$~(which describes the quality of zero-shot classifier in depicting its sampled distribution) and the distance between $Q'(\ry|\rvx)$ and $P(\ry|\rvx)$~(which describes the domain gap between the distribution of CLIP training dataset and the distribution of our in-question noisy dataset).

Specifically, we simply define the second part - \textit{domain gap}, as $\varepsilon_{domain}$ here, which represents how different the true prediction distribution~($Q'(\ry|\rvx)$) of CLIP training dataset is than the true prediction distribution of noisy dataset~($P(\ry|\rvx)$). This is technically irreducible but can be improved by making the CLIP training dataset more abundant and reducing its domain gap with the targeted classification dataset. 

On the other hand, the first part is further divided into two parts:
\begin{enumerate}
    \item the distance between $Q_{g,h}(\rvz|\rvx)$ and $Q'(\rvz|\rvx)$~(\textit{CLIP generalization error});
    \item the error induced by eq. (4) or eq. (9) when estimating \\ \noindent$P_{zeroshot}(\ry|\rvx=\vx)$ based on $Q'(\rvx, \rvz)$~(\textit{Prompt sampling error}).
\end{enumerate}
Intuitively, the \textit{CLIP generalization error} represents how well our CLIP model learns and generalizes, and the \textit{Prompt sampling error} represents how much extra bias we introduce when we try to approximate the integral with sampling.

\paragraph{CLIP generalization error}
Let us recall here the empirical risk on \textit{i.i.d} sampled dataset from the noisy CLIP distribution $Q$ with encoder $g\in\mathcal{G},h\in\mathcal{H}$ as $\hat{R}^Q(g, h)$:
\begin{equation*}
    \hat{R}^Q(g,h) = \frac{1}{M}\sum_{i=1}^M  L_{clip}(\vx_i, \vz_i; g,h),
\end{equation*}
the corresponding (unknown) empirical risk \textit{w.r.t} clean dataset as:
\begin{equation*}\label{eq:populationrisk}
    \hat{R}^{Q'}(g,h) = \frac{1}{M}\sum_{i=1}^M  L_{clip}(\vx_i, \vz'_i; g,h),
\end{equation*}
and the expected risk on the unknown clean CLIP distribution $Q'$ as $R^{Q'}(g,h)$, as: 
\begin{align*} 
\begin{split}
      R^{Q'}(g,h) &= E_{Q'} L_{clip}(\rvx, \rvz; g,h).
\end{split}
\end{align*}

Below we present how to bind the expected risk by empirical risk on a noisy CLIP training dataset. We denote $(\hat{g}, \hat{h}) = \arg\min_{g\in \mathcal{G}, h\in \mathcal{H}} \hat{R}^{Q}(g,h)$ as the empirical optimal model \textit{w.r.t} \textit{i.i.d} sampled dataset from $Q$, $(g^*,h^*) = \arg\min_{g\in \mathcal{G}, h \in \mathcal{H}} R^{Q'}(g,h)$ as the best-achievable model \textit{w.r.t} clean distribution $Q'$, and 

\noindent$(g_{bayes}, h_{bayes}) = \arg\min_{g,h} R^{Q'}(g,h)$ as the Bayes optimal model \textit{w.r.t} clean distribution $Q'$. We can decompose the excess risk of our learned empirical optimal model $\hat{f}$ over the Bayes optimal model $f_{bayes}$ as:
{\small
\begin{align}\label{eq:clip_riskdecomposition}
\begin{split}
R^{Q'}(\hat{g}, \hat{h}) - R^{Q'}(g_{bayes}, h_{bayes}) &= \underbrace{R^{Q'}(\hat{g}, \hat{h}) - R^{Q'}(g^*,h^*)}_{estimation\ error} \\
&+ \underbrace{R^{Q'}(g^*,h^*) - R^{Q'}(g_{bayes}, h_{bayes})}_{approximation\ error} \\
 &= R^{Q'}(\hat{g}, \hat{h}) - R^{Q'}(g^*,h^*) +\mathcal{B}_{approx} \\
 &\approx R^{Q'}(\hat{g}, \hat{h}) - R^{Q'}(g^*,h^*)
\end{split}
\end{align}
}
Exact analysis of approximation error is often intractable, we thus abbreviate it as $\mathcal{B}_{approx}$ and omit it in subsequent analysis. For estimation error, we have:
{\footnotesize
\begin{align}\label{eq:clip_estimation_riskdecomposition}
\begin{split}
R^{Q'}(\hat{g}, \hat{h}) - R^{Q'}(g^*,h^*) &= R^{Q'}(\hat{g}, \hat{h}) - \hat{R}^Q(\hat{g}, \hat{h}) + \hat{R}^Q(\hat{g}, \hat{h}) \\
 &- \hat{R}^Q(g^*,h^*) + \hat{R}^Q(g^*,h^*) - R^{Q'}(g^*,h^*) \\
 &\xRightarrow{\hat{R}^Q(\hat{g}, \hat{h}) - \hat{R}^Q(g^*,h^*) \leq 0} \\
 &\leq R^{Q'}(\hat{g}, \hat{h}) - \hat{R}^Q(\hat{g}, \hat{h}) + \hat{R}^Q(g^*,h^*) - R^{Q'}(g^*,h^*)\\
 &\leq 2 \mathtt{sup}_{g\in \mathcal{G}, h\in \mathcal{H}} |R^{Q'}(g,h) - \hat{R}^Q(g,h)| \\
\end{split}
\end{align}}

Supposing the range of $L_{clip}$ as $[0, l^{clip}_{\infty}]$ for all ($\rvx,\rvz$) in $\mathrm{sup}(Q)$ and $L_{clip}$ is $\lambda$-Lipschitz continuous \textit{w.r.t} $\rvz$, according to \Cref{theorem: rademacher} and triangle inequality, we have:
{\footnotesize
\begin{align}\label{eq:noisyrisk}
\begin{split}
    |R^{Q'}(g,h) - \hat{R}^Q(g,h)| &\leq  \overbrace{|R^{Q'}(g,h) - \hat{R}^{Q'}(g,h)|}^{\Cref{theorem: rademacher}} + \overbrace{|\hat{R}^{Q'}(g,h) - \hat{R}^Q(g,h)|}^{Lipschitz \ continuous} \\
    & \leq 2\mathfrak{R}(\mathcal{G}\circ\mathcal{H}) + l^{clip}_{\infty}\sqrt{\frac{\log(1/\delta)}{2M}} + \lambda \frac{1}{M}\sum_{i=1}^M \lVert \vz_i - \vz'_i \rVert_2 \\
    & \leq 2\mathfrak{R}(\mathcal{G}\circ\mathcal{H}) + l^{clip}_{\infty}\sqrt{\frac{\log(1/\delta)}{2M}} + \varepsilon_n
\end{split}
\end{align}
}
Here, we rewrite $\lambda \frac{1}{M}\sum_{i=1}^M \lVert \vz_i - \vz'_i \rVert_2$ as $\varepsilon_n$ which is the error term induced by the noisy descriptions is CLIP training set~($\vz_i \neq \vz'_i$). With \cref{eq:clip_riskdecomposition} and \cref{eq:noisyrisk}, we have:
{\small
\begin{equation}\label{eq:riskbound}
    R^{Q'}(\hat{g}, \hat{h}) - R^{Q'}(g_{bayes}, h_{bayes}) \leq 2(2\mathfrak{R}(\mathcal{G}\circ\mathcal{H}) + l^{clip}_{\infty}\sqrt{\frac{\log(1/\delta)}{2M}} + \varepsilon_n)
\end{equation}
}
To further connect the generalization error bound above and the distance of estimated probability $Q_{g,h}(\rvz|\rvx)$ and $Q'(\vz|\vx)$, we have:
\begin{align}
\begin{split}
        R^{Q'}(g,h) &= E_{Q'} L_{clip}(\rvx, \rvz; g,h) \\
        &=-\frac{1}{2}\int Q'(\rvx) \int Q'(\rvz|\rvx) \log Q_{g,h}(\rvz|\rvx) d\rvz d\rvx \\
        &-\frac{1}{2}\int Q'(\rvz) \int Q'(\rvx|\rvz) \log Q_{g,h}(\rvx|\rvz) d\rvx d\rvz \\
        &=\frac{1}{2}\int Q'(\rvx) D_{KL}(Q'(\rvz|\rvx), Q_{g,h}(\rvz|\rvx)) d\rvx \\
        &-\frac{1}{2}\int Q'(\rvx) \int Q'(\rvz|\rvx) \log Q'(\rvz|\rvx) d\rvz d\rvx \\
        &-\frac{1}{2}\int Q'(\rvz) \int Q'(\rvx|\rvz) \log Q_{g,h}(\rvx|\rvz) d\rvx d\rvz \\
        &\geq \frac{1}{2}\int Q'(\rvx) D_{KL}(Q'(\rvz|\rvx), Q_{g,h}(\rvz|\rvx)) d\rvx 
\end{split}
\end{align}
We define the distribution distance as:
\begin{equation}\label{eq:distance}
    d(Q_{g,h}(\rvz|\rvx), Q'(\rvz|\rvx)) \triangleq \frac{1}{2}\int Q'(\rvx) D_{KL}(Q'(\rvz|\rvx), Q_{g,h}(\rvz|\rvx)) d\rvx.
\end{equation}

Intuitively, when and only when the learned model is Bayes optimal, we have a zero distance between the estimated probability and the ground-truth probability. According to \cref{eq:riskbound}, we thus have:
{\small
\begin{align}\label{eq:distance_clip}
\begin{split}
    d(Q_{\hat{g}, \hat{h}}(\rvz|\rvx), Q'(\rvz|\rvx)) 
    &\leq R^{Q'}(g_{bayes}, h_{bayes}) +2(2\mathfrak{R}(\mathcal{G}\circ\mathcal{H})  \\&+l^{clip}_{\infty}\sqrt{\frac{\log(1/\delta)}{2M}} + \varepsilon_n) \\
    &\leq \varepsilon_{clip}+ 2(2\mathfrak{R}(\mathcal{G}\circ\mathcal{H}) + l^{clip}_{\infty}\sqrt{\frac{\log(1/\delta)}{2M}} + \varepsilon_n)
\end{split}
\end{align}
}
Here, $\varepsilon_{clip}$ denotes the expected risk of the Bayes optimal CLIP model.

\paragraph{Prompt sampling and designing}
We then take step two into consideration. According to eq. (3), with $Q_{\hat{g},\hat{h}}(\rvz|\rvx)$ we can estimate $P_{zeroshot}(\ry|\rvx)$. To quantify the additional error of the sampling process~eq. (4) or eq. (9), we denote as $\Delta$ an error coefficient which represents how much extra error has been induced. Let us recall the domain gap~($\varepsilon_{domain}$) before, we thus have:
\begin{theorem}[\textsc{Estimation with zero-shot classifier}]
Let $\mathcal{G}, \mathcal{H}$ be the hypothesis space of vision encoder $g$ and language encoder $h$. Let us denote the rademacher complexity as $\mathfrak{R}(\mathcal{G}\circ\mathcal{H})$ of the combined CLIP model. Supposing the range of $L$ from eq. (2) as $[0, l^{clip}_{\infty}]$ for all ($\vx,\vz$) in $\mathrm{sup}(Q)$ with $g,h \in \mathcal{G}, \mathcal{H}$. Then, for any $\delta >0$, with probability at least $1-\delta$ we have the following holds:
{\footnotesize
\begin{equation*}
d(P_{zeroshot}, P) \leq \varepsilon_{domain} + \Delta\textcolor{gray}{(\lambda_0\varepsilon_{clip} + \lambda_1 \mathfrak{R}(\mathcal{G}\circ\mathcal{H}) + \lambda_2 l^{clip}_{\infty}\sqrt{\frac{\log 1/\delta}{M}} + \lambda_3\varepsilon_{n})}
\end{equation*}
}
with $\lambda_0, \lambda_1, \lambda_2, \lambda_3 > 0$. Here, $\varepsilon_{domain}$ denotes the bias term induced by the domain gap between $Q$ and $P^{true}$, $\varepsilon_{clip}$ denotes the expected risk of the Bayes optimal CLIP model, and $\Delta \geq 1$ denotes the bias coefficient induced by designing prompts and sampling in eq. (3).
\end{theorem}

\subsection{Derivation of \textsc{Theorem 3.2}}
The derivation of \textsc{Theorem 3.2} follows a similar but rather simpler process. Denoting $f'\in\mathcal{F}$ as the induced classifier, we have correspondingly the empirical risk with the noisy training set:
\begin{equation*}
    \hat{R}^{P^n}(f') = \frac{1}{N}\sum_{i=1}^N  L_{noisy}(\vx_i,y_i; f'),
\end{equation*}
the corresponding (unknown) empirical risk \textit{w.r.t} clean dataset as:
\begin{equation*}\label{eq:populationrisk2}
    \hat{R}^P(f') = \frac{1}{N}\sum_{i=1}^N  L_{noisy}(\vx_i,y'_i; f'),
\end{equation*}
and the expected risk as $R^P(f)$, as: 
\begin{align*} 
\begin{split}
      R^{P}(f') &= E_{P} L_{noisy}(\rvx,\ry; f').
\end{split}
\end{align*}

Specifically, with $Q, Q', \vz_i, \vz'_i, M$ replaced by $P, P', y_i, y'_i, N$, similar to \cref{eq:riskbound}, we have:
\begin{equation}\label{eq:noisy_riskbound}
    R^P(\hat{f'}) - R^P(f'_{bayes}) \leq 2(2\mathfrak{R}(\mathcal{F}) + l^{noisy}_{\infty}\sqrt{\frac{\log(1/\delta)}{2N}}) + \varepsilon_{noise}
\end{equation}
Here, $\varepsilon_{noise}$ denotes the difference term induced by the difference between the noisy distribution $P^n(\ry|\rvx)$ and clean distribution $P(\ry|\rvx)$, practically, the label noise in the training dataset. 

To similarly connect the generalization error bound above and the distance of estimated probability $P_f'(\ry|\vx)$ and $P(\ry|\vx)$, with $L_{noisy}$ as the cross-entropy loss, we have:
\begin{align}
\begin{split}
        R^P(f') &= E_{P} L_{noisy}(\vx, \ry; f') \\
        &=-\int P(\vx) \int P(\ry|\vx) \log P_{f'}(\ry|\vx) dy d\vx \\
        &= \int P(\vx) D_{KL}(P(\ry|\vx), P_{f'}(\ry|\vx)) d\vx \\
        &- \int P(\vx) \int P(\ry|\vx) \log P(y|\vx) dy d\vx \\
        &\geq \int P(\vx) D_{KL}(P(\ry|\vx), P_{f'}(\ry|\vx)) d\vx
\end{split}
\end{align}
Similar to \cref{eq:distance}, we define:

\begin{equation}
    d(P_{f'}(\ry|\vx), P(\ry|\vx)) = \frac{1}{2}\int P(\vx) D_{KL}(P(\ry|\vx), P_{f'}(\ry|\vx)) d\vx
\end{equation}
Let us denote as $\varepsilon_{induced} \triangleq R^P(f_{bayes})$ the expected risk of the Bayes optimal induced classifier, we then have:
\begin{theorem}[\textsc{Estimation with induced classifier}]
Let $\mathcal{F}$ be the hypothesis space of induced classifier $f'$. Let us denote the Rademacher complexity as $\mathfrak{R}(\mathcal{F})$ of the induced classifier. Supposing the range of $L$ for training $f'$ as $[0, l^{noisy}_{\infty}]$ for all ($\vx,y$) in $\mathrm{sup}(P)$ with $f'\in\mathcal{F}$. Then, for any $\delta >0$, with probability at least $1-\delta$ we have the following holds:
\begin{equation*}
d(P_{induced}, P) \leq \varepsilon_{noise} + \textcolor{gray}{\lambda_0 \varepsilon_{induced} + \lambda_1 \mathfrak{R}(\mathcal{F}) + \lambda_2 l^{noisy}_{\infty}\sqrt{\frac{\log 1/\delta}{N}}}
\end{equation*}
with $\lambda_0, \lambda_1, \lambda_2 > 0$. Here, $\varepsilon_{noise}$ denotes the difference term induced by the label noise in the training dataset, and $\varepsilon_{induced}$ denotes the expected risk of the Bayes optimal induced classifier. 
\end{theorem}

\section{Experiment details}
\label{sup:a}
\subsection{Dataset details}
We conduct extensive experiments on two standard benchmarks with synthetic label noise, CIFAR-10 and CIFAR-100, and four real-world noisy datasets, Red Mini-ImageNet, WebVision, ANIMAL-10N, and Clothing1M. 

\textbf{CIFAR10} and \textbf{CIFAR100} datasets comprise 50,000 images. Following established conventions, we assess our method's performance with two types of artificial noise:``symmetric noise," wherein labels are randomly flipped across all samples using a uniform distribution, and ``asymmetric noise," wherein labels of visually similar categories, such as Horse $\leftrightarrow$ Deer and Dog $\leftrightarrow$ Cat, are randomly interchanged. Moreover, we conduct experiments with various noise levels: 20\%, 50\%, 80\% and 90\% symmetric noise, as well as 40\% asymmetric noise, adhering to the settings in DivideMix~(\cite{dividemix}). For instance-dependent noise, we utilize the label noise file provided by \cite{chen2021INDnoise}.

\textbf{Red Mini-ImageNet} dataset~\cite{mentormix} is a real-world dataset containing a total of 100 categories. It is an extension of the Mini-Imagenet dataset, where noise is introduced at varying ratios. Specifically, noisy images and their respective labels are obtained by crawling the internet, and these noisy images replace the original images in the Mini-ImageNet dataset, with different noise ratios. To ensure a fair comparison with previous studies~\cite{instancegm,famus}, the images are resized from their original size of 84×84 pixels to 32×32 pixels. Moreover, in accordance with the existing literature~\cite{instancegm,famus}, we utilize noise ratios of 20\%, 40\%, 60\%, and 80\%.

\textbf{WebVision}~\cite{webvision} is an extensive dataset comprising 1,000 classes of images obtained through web crawling. In line with previous studies~\cite{mentornet,dividemix,moit}, we evaluate our methods using the top 50 classes from the Google Subset of WebVision. The estimated noise ratio for this subset is approximately 20\%.

\textbf{ANIMAL-10N}~\cite{selfie} is a recently introduced real-world noisy dataset comprises 10 classes of animals. The dataset has undergone manual labeling, with an estimated label noise ratio of around 8\%. Similar to the CIFAR datasets, ANIMAL-10N consists of 50,000 training images and 10,000 test images.

\textbf{Clothing1M}~\cite{clothing1mdataset} is a large-scale dataset containing 14 classes of clothing images, obtained by crawling online shopping websites. It consists of a substantial collection of 1 million noisy images. The estimated noise ratio for this dataset is approximately 38.5\%.

\subsection{Implementation details}
\label{sup:b}
We use the CLIP model with VIT-B/32 backbone in all experiments except for specific ablations. In all experiments, our default approach is \textit{CLIPCleaner + MixFix}~(\textbf{Ours}). By default, we train the network with an SGD optimizer with a momentum of 0.9 in all experiments.

For \textbf{CIFAR10} and \textbf{CIFAR100}, we use a PresActResNet-18~\cite{preactresnet} as the backbone in all experiments following previous works. For CIFAR10, we set $\theta_{loss} = 0.5, \theta_{consistency} = 0.8$ for \textit{CLIPCleaner} and $\theta_r = 0.8, \theta'_r = 0.9$ for \textit{MixFix}; For CIFAR10, we set $\theta_{loss} = 0.5, \theta_{consistency} = 0.8$ for \textit{CLIPCleaner} and $\theta_r = 0.7, \theta'_r = 0.8$ for \textit{MixFix}. We train both networks for 300 epochs with a weight decay of 5e-4. The initial learning rate is 0.02 and is controlled by a cosine annealing scheduler. The batchsize is fixed as 128. 

For \textbf{Red Mini-ImageNet}, we also use a PresActResNet-18~\cite{preactresnet} as the backbone following previous works~\cite{instancegm,famus}. For \textit{CLIPCleaner}, we set $\theta_{loss} = 0.5, \theta_{consistency} = 0.8$. For \textit{MixFix}, we set $\theta_r = 0.8, \theta'_r = 0.95$. We train the network for 300 epochs with a weight decay of 5e-4. The initial learning rate is 0.02 and reduced by a factor of 10 after 200 and 250 epochs. The batchsize is fixed as 64. 

For \textbf{WebVision}, we use InceptionResNetv2 as the backbone following~\cite{dividemix}. For \textit{CLIPCleaner}, we set $\theta_{loss} = 0.5, \theta_{consistency} = 1$. For \textit{MixFix}, we set $\theta_r = 0.7, \theta'_r = 1.0$. 
We train the network for 150 epochs with a weight decay of 1e-4. The initial learning rate is 0.01 and reduced by a factor of 10 after 80 and 120 epochs. The batchsize is fixed as 32. 

For \textbf{Clothing1M}, we use a ResNet50 as the backbone following~\cite{dividemix} with ImageNet pretrained weights. For \textit{CLIPCleaner}, we set $\theta_{loss} = 0, \theta_{consistency} = 0.5$. For \textit{MixFix}, we set $\theta_r = 0.7, \theta'_r = 1.0$. We train the network for 150 epochs with a weight decay of 1e-3. The initial learning rate is 0.002 and reduced by a factor of 10 after 50 and 100 epochs. The batchsize is fixed as 32. 

For \textbf{ANIMAL-10N}, we use a VGG-19~\cite{vgg19} as the backbone with batch-normalization following~\cite{selfie}. For \textit{CLIPCleaner}, we set $\theta_{loss} = 0.5, \theta_{consistency} = 0.8$. For \textit{MixFix}, we set $\theta_r = 0.7, \theta'_r = 0.99$. We train the network with an SGD optimizer for 300 epochs with a momentum of 0.9 and weight decay of 5e-4. The initial learning rate is 0.02 and reduced by a factor of 10 after 150 and 250 epochs. The batchsize is fixed as 128. 

\section{Intuition of sample selection with fine-grained dataset}
\label{appdix:sampleselection}
For better clarification, we here start with a more detailed problem formulation. Given a dataset of training samples ${(\vx_i,y_i)}_{i=1}^{N}$ \textit{i.i.d} sampled from a noisy joint distribution $P^n(\rvx,\ry)$ with $supp(P) = \{\rvx \in R^{C\times H\times W}, \ry\in \{1,..., K\}\}$ and $K$ denotes the number of semantic classes, the goal of supervised learning is to learn a model $f$ that can accurately predict the true labels $\ry$ for new, unseen examples. Mathematically, we often optimize the empirical risk with samples \textit{i.i.d} sampled from the noisy distribution:
\begin{equation*}
\hat{R}^{P^n}(f) = \frac{1}{N}\sum_{i=1}^N L(\vx_i,y_i;f)
\end{equation*}
Here $L$ can be any applicable classification-calibrated surrogate loss to \textit{0-1 loss}, normally we use \textit{Cross-Entropy loss}:
\begin{equation*}
    L(\vx_i,y_i;f) = -\log \frac{\exp(f(\vx_i)_{y_i})}{\sum_{j=1}^K \exp(f(\vx_i)_j)}.
\end{equation*}
Owing to the \textit{ERM principle}, we can uniformly minimize \textit{w.r.t} the expected risk by minimizing above empirical risk:
\begin{equation*}
R^{P^n}(f) = E_{P^n(\rvx, \ry)}L(\vx,y;f)
\end{equation*}

However, in this work we focus on learning with noisy labels, that is to say, there exists a discrepancy between the noisy training distribution $P^n(\vx,y)$ and clean unknown distribution $P(\vx,y)$. In this condition, for the same specific model $f$, we have the expected risk on the real distribution as:
\begin{equation*}
R^P(f) = E_{P(\rvx, \ry)}L(\vx,y;f)
\end{equation*}
To bridge the distribution discrepancy, we can easily find that:
\begin{equation*}
    R^P(f) = E_{P(\rvx, \ry)}L(\vx,y;f) = E_{P^n(\rvx, \ry)}\frac{P(\rvx=\vx, \ry=y)}{P^n(\rvx=\vx, \ry=y)} L(\vx,y;f).
\end{equation*}

The above equation widely applies in various problems involving distribution mismatch between training and testing phases, such as class imbalance.
In particular, in the interested LNL problem setting, we often assume $\frac{P(\rvx, \ry)}{P^n(\rvx, \ry)} = \frac{P(\ry|\rvx)P(\rvx)}{P^n(\ry|\rvx)P^n(\rvx)}= \frac{P(\ry|\rvx)}{P^n(\ry|\rvx)}$ as label noise normally does not affect the sample itself~($P(\rvx) = P^n(\rvx)$). We then get the corresponding weighted empirical risk with noisy labels,
\begin{align}\label{eq:erm_weighted}
\hat{R}^P(f) &= \frac{1}{N}\sum_{i=1}^N \frac{P(\ry=y_i|\rvx=\vx_i)}{P^n(\ry=y_i|\rvx=\vx_i)} L(\vx_i,y_i;f)\\
&= \frac{1}{N}\sum_{i=1}^N w_i L(\vx_i,y_i;f).
\end{align}
Please note, that similar derivation has already been discussed in \cite{sample_reweighting}. Obviously, with optimal weights~($w_i =\frac{P(\ry=y_i|\rvx=\vx_i)}{P^n(\ry=y_i|\rvx=\vx_i)}$) we can ensure a risk-consistent classifier \textit{w.r.t} clean distribution with even noisy labels. 
Since $P(\ry=y_i|\rvx=\vx_i)$ and $P^n(\ry=y_i|\rvx=\vx_i)$ are typically both unknown for $\vx_i$, the objective of sample selection methods often revolves around estimating these two to roughly estimate the optimal weights. 
Specifically, the noisy label $y_i$ can serve as a confident proxy of the noisy distribution $P^n(\ry=y_i|\rvx=\vx_i)$, making our focus on utilizing an additional auxiliary classifier $\Tilde{P}(\ry=y_i|\rvx=\vx_i)$ to estimate $P(\ry=y_i|\rvx=\vx_i)$. Then, a manually-designed sample selection strategy can be utilized to further refine the estimation.

\textit{In most conditions, we can restrict the weights as binary since for most samples in most classification datasets, $P(\ry|\rvx=\vx_i)$ tends to be highly centred around only one class - this leads to $w_i \approx 1$ or $w_i \approx 0$. However, in scenarios such as a fine-grained dataset, $w_i$ tends to be less extreme, and a binary sample selection may be sub-optimal. We leave this to further exploration.}

\end{document}